%% file: main.tex
\documentclass{article}

\PassOptionsToPackage{numbers,sort&compress}{natbib}
\PassOptionsToPackage{font=small}{caption}
\usepackage[final]{corl_2018}

\usepackage{amsmath}
\usepackage{algpseudocode}
\usepackage{subcaption}
\usepackage{graphicx}
\usepackage{xcolor}
\usepackage{wrapfig}
\usepackage{titlesec}
\usepackage{setspace}

\AtBeginDocument{%
}
\title{SPNets: Differentiable Fluid Dynamics \\for Deep Neural Networks\\\vspace{-0.0cm}}

\author{
  Connor Schenck\\
  University of Washington, Nvidia\\
  \texttt{schenckc@cs.washington.edu} \\
  \And
  Dieter Fox\\
  Nvidia, University of Washington\\
  \texttt{dieterf@nvidia.com} \\
}

\begin{document}
\maketitle
% \vspace{-1.0cm}
\begin{abstract}
In this paper we introduce Smooth Particle Networks (SPNets), a framework for integrating fluid dynamics with deep networks.
SPNets adds two new layers to the neural network toolbox: ConvSP and ConvSDF, which enable computing physical interactions with unordered particle sets.
We use these layers in combination with standard neural network layers to directly implement fluid dynamics inside a deep network, where the parameters of the network are the fluid parameters themselves (e.g., viscosity, cohesion, etc.).
Because SPNets are implemented as a neural network, the resulting fluid dynamics are fully differentiable.
We then show how this can be successfully used to learn fluid parameters from data, perform liquid control tasks, and learn policies to manipulate liquids.
\end{abstract}


\section{Introduction}

% Motivate
From mixing dough to cleaning automobiles to pouring beer, liquids are an integral part of many everyday tasks.
Humans have developed the skills to easily manipulate liquids in order to solve these tasks, however robots have yet to master them.
While recent results in deep learning have shown a lot of progress in applying deep neural networks to challenging robotics tasks involving rigid objects~\cite{byravan2018,srinivas2018,wahlstrom2015}, there has been relatively little work applying these techniques to liquids.
One major obstacle to doing so is the highly unstructured nature of liquids, making it difficult to both interface the liquid state with a deep network and to learn about liquids completely from scratch.

% Solution
In this paper we propose to combine the structure of analytical fluid dynamics models with the tools of deep neural networks to enable robots to interact with liquids.
Specifically, we propose Smooth Particle Networks (SPNets), which adds two new layers, the ConvSP layer and the ConvSDF layer, to the deep learning toolbox.
These layers allow networks to interface directly with unordered sets of particles.
We then show how we can use these two new layers, along with standard layers, to directly implement fluid dynamics using Position Based Fluids (PBF) \citep{macklin2013} inside the network, where the parameters of the network are the fluid parameters themselves (e.g., viscosity or cohesion).
Because we implement fluid dynamics as a neural network, this allows us to compute full analytical gradients.
We evaluate our fully differentiable fluid model in the form of a deep neural network on the tasks of learning fluid parameters from data, manipulating liquids, and learning a policy to manipulate liquids.
% We show that our model can be used to enable robots to solve a variety of tasks involving liquids.
In this paper we make the following contributions 1) a fluid dynamics model that can interface directly with neural networks and is fully differentiable, 2) a method for learning fluid parameters from data using this model, and 3) a method for using this model to manipulate liquid by specifying its target state rather than through auxiliary functions.
In the following sections, we discuss related work, the PBF algorithm, SPNets, and our evaluations of our method.

% Outline

% The rest of this paper is as follows. 
% The next section discusses related work.
% Section~\ref{sec:pbf} details the fluid dynamics algorithm PBF.
% Section~\ref{sec:spnets} describes SPNets and how we implement PBF.
% Section~\ref{sec:eval} describes the tasks we evaluated our model on and the results.
% Finally, section~\ref{sec:conclusion} concludes the paper and lays out avenues for future work.

\section{Related Work}

% Liquids and robots
% -*yamaguchi cmu stuff
% -*kunze sim planning stuff
% -*My stuff
% -*A probabilisty approach to liquid level detection in cups (do2016)
% -*Discriminating liquids using a robotic kitchen assistent (elbrechter2015)
% -Trajectory planning fo rmeal assist robot considering spilling avoidance
% -Motion planning for fluid manipulation using simplified dynamics
% -feedback motion planning for liquid pouring using supervised learning
% -*seeing liquid from static snapshots (paulun2015)
% -*Daytime water detection based on color variation/Daytime water detection based on sky reflections (rankin*)
% -XIncorporating failure-to-success transitions in imitation learning for a dynamic pouring task (langsfeld2014)
% -XLearning to pour with a robot arm combining goal and shape learning for dynamic movement primitives (tamosiunaite2011)
% -XDesigning robot learners that ask good questions (cakmak2012)
% -Force-based robot learning of pouring skills using parametric hidden markov models
% -Precise dispensing of liquids using visual feedback
Liquid manipulation is an emerging area of robotics research.
In recent years, there has been much research on robotic pouring~\citep{schenck2016c,yamaguchi2016c,kuriyama2008,pan2017,rozo2013,kennedy2017,langsfeld2014,tamosiunaite2011,cakmak2012}.
There have also been several papers examining perception of liquids~\citep{yamaguchi2016,schenckc2018a,do2016,elbrechter2015,paulun2015,rankin2011}.
Some work has used simulators to either predict the effects of actions involving liquids~\citep{kunze2015}, or to track and reason about real liquids~\citep{schenckc2017a}.
However, all of these used either task specific models or coarse fluid dynamics, with the exception of \citep{schenckc2017a}, which used a liquid simulator, although it was not differentiable.
Here we propose a fluid dynamics model that is fully differentiable and show how to use it solve several tasks.

% Learning about liquids/interfacing deep nets with points
% -Discriminating liquids using a robotic kitchen assistent
% -Adapatable pouring teaching robots not to spill using fast but approximate fluid simulation
% -Polyfes regression forest stuff
% -tompson fluid dynamics stuff
% -Latent-space physics: towards learning the temporal evolution of fluid flow
% -PointNet
% -Vote3Deep
% -SplatNet
One task we evaluate is learning fluid parameters (e.g., viscosity or cohesion) from data.
Work by Elbrechter {\it et al.} \citep{elbrechter2015} and Guevara {\it et al.} \citep{guevara2018} focused on learning fluid parameters using actions to estimate differences between the model and the data.
Other work has focused on learning fluid dynamics via hand-crafted features and regression forests~\citep{ladicky2015}, via latent-state physics models~\citep{wiewel2018}, or via conventional simulators combined with a deep net trained to solve the incompressibility constraints~\citep{tompson2016}.
Both \citep{wiewel2018} and \citep{tompson2016} use grid-based fluid representations, which allows them to use standard 3D convolutions to implement their deep learning models.
Both \citep{miyanawala2017} and \citep{baymani2015} also used standard convolutions to implement fluid physics using neural networks.
In this paper, however, we use a particle-based fluid representation due to its greater efficiency for sparse fluids. 
In \citep{ladicky2015} the authors also use a particle-based representation, however they require hand-crafted features to allow their model to compute particle-particle interactions.
Instead, we directly interface the particles with the model.
While there have been several recent papers that develop methods for interfacing unordered point sets with deep networks~\citep{qi2017,engelcke2017,su2018}, these methods focus on the task of object recognition, a task with significantly different computational properties than fluid dynamics.
For that reason, we implement new layers for interfacing our model with unordered particle sets.

%miyanawala2017, baymani2015, 

% Differentiable physics/rendering/fluid dynamics
% -Macklin stuff
% -SPH stuff/Navier stokes/general fluid dynamics
% -A differentiable physics engine for deep learning http://phys.csail.mit.edu/papers/1.pdf, longer version https://arxiv.org/pdf/1611.01652.pdf
% -differentiable renderer stuff
The standard method of solving the Navier-Stokes equations~\cite{acheson1990} for computing fluid dynamics using particles is Smoothed Particle Hydrodynamics (SPH)~\cite{gingold1977}.
In this paper, however, we use Position Based Fluids (PBF)~\citep{macklin2013} which was developed as a counterpart to SPH.
SPH computes fluid dynamics for {\it compressible} fluids (e.g., air); PBF computes fluid dynamics for {\it incompressible} fluids (e.g., water).
Additionally, our model is differentiable with analytical gradients.
There has been some work in robotics utilizing differentiable physics models~\citep{degrave2016} as well as differentiable rendering~\citep{loper2014}.
There has also been work on learning physics models using deep networks such as Interaction Networks~\citep{battaglia2016,watters2017}, which model the interactions between objects as relations, and thus are also fully differentiable.
However, these works were primarily focused on simulating rigid body physics and physical forces such as gravity, magnetism, and springs.
To the best of our knowledge, our model is the first fully differentiable particle-based fluid model.

\section{Position Based Fluids}
\label{sec:pbf}

\begin{wrapfigure}[17]{R}{7.4cm}
\vspace{-1.0cm}
\begin{algorithmic}[1]
\Function{UpdateFluid}{P, V}
    \State $V' = \Call{ApplyForces}{V}$ \label{line:applyForces}
    \State $P' = P + \frac{V'}{{\Delta}t}$ \label{line:applyVelocities}
    \While{$\neg\Call{ConstraintsSatisfied}{P'}$}
        \State ${\Delta}P^\omega = \Call{SolvePressure}{P'}$ \label{line:solvePressure}
        \State ${\Delta}P^c = \Call{SolveCohesion}{P'}$ \label{line:solveCohesion}
        \State ${\Delta}P^s = \Call{SolveSurfaceTension}{P'}$ \label{line:solveTension}
        \State $P' = P' + {\Delta}P^\omega + {\Delta}P^c + {\Delta}P^s$
        \State $P' = \Call{SolveObjectCollisions}{P'}$ \label{line:solveCollisions}
    \EndWhile
    \State $V' = \frac{P' - P}{{\Delta}T}$ \label{line:newvel}
    \State $V' = V' + \Call{ApplyViscosity}{P', V'}$ \label{line:viscosity}
    \State\Return $P', V'$
\EndFunction
\end{algorithmic}
\caption{The PBF algorithm. $P$ is the list of particle locations, $V$ is the list of particle velocities, and ${\Delta}T$ is the timestep duration.}
\label{fig:pbf}
\end{wrapfigure}

In this paper, we implement fluid dynamics using Position Based Fluids (PBF) \cite{macklin2013}.
PBF is a Lagrangian approximation of the Navier-Stokes equations for incompressible fluids \cite{acheson1990}.
That is, PBF uses a large collection of particles to represent incompressible fluids such as water, where each particle can be thought of as an individual ``packet'' of fluid.
We chose a particle-based representation for our fluids rather than a grid-based (Eulerian) representation as for sparse fluids, particles have better resolution for fewer computational resources.
We briefly describe PBF here and direct the reader to \citep{macklin2013} for details.

Figure~\ref{fig:pbf} shows a general outline of the PBF algorithm for a single timestep.
First, at each timestep, external forces are applied to the particles (lines~\ref{line:applyForces}--\ref{line:applyVelocities}), then particles are moved to solve the constraints (lines~\ref{line:solvePressure}--\ref{line:solveCollisions}), and finally the viscosity is applied (line~\ref{line:viscosity}), resulting in new positions $P'$ and velocities $V'$ for the particles.
In this paper, we consider three constraints on our fluids: pressure, cohesion, and surface tension, which correspond to the three inner loop functions in Figure~\ref{fig:pbf} (lines~\ref{line:solvePressure}--\ref{line:solveTension}).
%\textproc{SolvePressure} (line~\ref{line:solvePressure}), \textproc{SolveCohesion} (line~\ref{line:solveCohesion}), and \textproc{SolveSurfaceTension} (line~\ref{line:solveTension}).
Each computes a ${\delta}p_i$ correction in position for each particle $i$ that best satisfies the given constraint.

% The pressure correction ${\delta}p^{\omega}_i$ for each particle $i$ is computed as follows
% \begin{equation}
%     {\delta}p^{\omega}_i = \displaystyle\sum_{j \in P-\{i\}} n_{ji}(\omega_i + \omega_j)W_\omega(d_{ij}, h)  \label{eq:pressureCorrection} 
% \end{equation}
% where $n_{ji}$ is the normal vector from particle $j$ to particle $i$, $\omega_k$ is the pressure at particle $k$, $W_\omega$ is a kernel function, $d_{ij}$ is the distance from $i$ to $j$, and $h$ is the cutoff for $W_\omega$ (that is, for all particles further than $h$ apart, $W_\omega$ is 0).
% The pressure $\omega_k$ at every particle $k$ is computed as
% \begin{equation}
%     \omega_k = \lambda_\omega \max\left( \rho_k - \rho_0, 0 \right) \label{eq:pressure}
% \end{equation}
% where $\lambda_\omega$ is the pressure constant, $\rho_k$ is the density of the fluid at particle $k$, and $\rho_0$ is the rest density of the fluid.
% The density $\rho_k$ is computed the same in PBF and SPH as follows
% \begin{equation}
%     \rho_k = \displaystyle\sum_{j \in P} m_j W_\rho(d_{kj}, h) \label{eq:density}
% \end{equation}
% where $m_j$ is the mass of particle $j$ and $W_\rho$ is a kernel function.

The pressure correction ${\delta}p^{\omega}_i$ for each particle $i$ is computed to satisfy the constant pressure constraint.
Intuitively, the pressure correction step finds particles with pressure higher than the constraint allows (i.e., particles where the density is greater than the ambient density), then moves them along a vector away from other high pressure particles, thus reducing the overall pressure and satisfying the constraint.
The pressure correction ${\delta}p^{\omega}_i$ for each particle $i$ is computed as
\begin{equation}
    \scalebox{0.8}{${\delta}p^{\omega}_i = \displaystyle\sum_{j \in P-\{i\}} n_{ji}(\omega_i + \omega_j)W_\omega(d_{ij}, h)$}  \label{eq:pressureCorrection}
\end{equation}
where $n_{ji}$ is the normalized vector from particle $j$ to particle $i$, $\omega_k$ is the pressure at particle $k$, $W_\omega$ is a kernel function (i.e., monotonically decreasing continuous function), $d_{ij}$ is the distance from $i$ to $j$, and $h$ is the cutoff for $W_\omega$ (that is, for all particles further than $h$ apart, $W_\omega$ is 0).
The pressure at each particle is computed as
\begin{equation}
     \scalebox{0.8}{$\omega_k = \lambda_\omega \max\left( \rho_k - \rho_0, 0 \right)$} \label{eq:pressure} 
\end{equation}
where $\lambda_\omega$ is the pressure constant, $\rho_k$ is the density of the fluid at particle $k$, and $\rho_0$ is the rest density of the fluid. 
Density at each particle is computed as
\begin{equation}
     \scalebox{0.8}{$\rho_k = \displaystyle\sum_{j \in P} m_j W_\rho(d_{kj}, h)$} \label{eq:density} 
\end{equation}
where $m_j$ is the mass of particle $j$.
For $W_\omega$ we use \scalebox{0.8}{$\frac{30}{\pi h^3} \left( 1 - \frac{d}{h} \right) \frac{1}{h}$} and for \scalebox{0.8}{$W_\rho$ we use $\frac{15}{\pi h^3} \left( 1 - \frac{d}{h} \right)^2$}, the same as used in~\citep{macklin2013}.
The details for computing \textproc{SolveCohesion}, \textproc{SolveSurfaceTension}, and \textproc{ApplyViscosity} are described in the appendix.

To compute the next set of particle locations $P'$ and velocities $V'$ from the current set $P, V$, these functions are applied as described in the equation in figure~\ref{fig:pbf}.
For the experiments in this paper, the constants are empirically determined and we set $h$ to $0.1$.

\section{Smooth Particle Networks}
\label{sec:spnets}

In this paper, we wish to implement Position Based Fluids (PBF) with a deep neural network.
%However current standard neural networks lack the functionality to implement PBF inside the network structure.
%While \citep{qi2017} developed PointNet layers which work on unordered point sets, these were designed for object recognition and segmentation and are not well suited for fluid dynamics.
%Thus, in order to implement PBF in a network, we need to add new functionality.
%Specifically, we propose two new neural network layers for handling unordered sets of particles.
Current networks lack the functionality to interface with unordered sets of particles, so we propose two new layers..
The first is the ConvSP layer, which computes particle-particle pairwise interactions, and the second is the ConvSDF layer, which computes particle-static object interactions\footnote{The code for SPNets is available at \url{https://github.com/cschenck/SmoothParticleNets}}.
We combine these two layers with standard operators (e.g., elementwise addition) to reproduce the algorithm in figure~\ref{fig:pbf} inside a deep network.
%Since we exactly reproduce PBF inside the network, the parameters are the $\lambda_*$ constants descried in section~\ref{sec:pbf}.
The parameters are the $\lambda_*$ values descried in section~\ref{sec:pbf}.
We implemented both forward and backward functions for our layers in PyTorch~\cite{pytorch} with graphics processor support.
%We implemented both our layers with support for GPUs to enable efficient processing times and with native support for analytical gradients.
%We used PyTorch~\cite{pytorch} to construct the fluid dynamics by combining our layers with standard network layers, which also enables us to take advantage of its automatic differentiation for our PBF implementation.

\subsection{ConvSP}

The ConvSP layer is designed to compute particle to particle interactions.
To do this, we implement the layer as a smoothing kernel over the set of particles.
That is, ConvSP computes the following 
\[ \scalebox{0.8}{$ConvSP(X, Y) = \Bigg\{ \displaystyle\sum_{j \in X} y_j W(d_{ij}, h) \;\Big|\; i \in X \Bigg\}$} \]
where $X$ is the set of particle locations and $Y$ is a corresponding set of feature vectors\footnote{In general, these features can represent any arbitrary value, however for the purposes of this paper, we use them to represent physical properties of the particles, e.g., mass or density.}, $y_j$ is the feature vector in $Y$ associated with $j$, $W$ is a kernel function, $d_{ij}$ is the distance between particles $i$ and $j$, and $h$ is the cutoff radius (i.e., for all $d_{ij} > h$, $W(d_{ij}, h) = 0$).
This function computes the smoothed values over $Y$ for each particle using $W$.

While this function is relatively simple, it is enough to enable the network to compute the solutions for pressure, cohesion, surface tension, and viscosity (lines \ref{line:solvePressure}--\ref{line:solveTension} and \ref{line:viscosity} in figure~\ref{fig:pbf}).
In the following paragraphs we will describe how to compute the pressure solution using the ConvSP layer.
Computing the other 3 solutions is nearly identical.

To compute the pressure correction solution in equation~\eqref{eq:pressureCorrection} above, we must first compute the density $\rho_k$ at each particle $k$.
Equation~\eqref{eq:density} describes how to compute the density.
This equation closely matches the ConvSP equation from above.
To compute the density at each particle, we can simply call $ConvSP(P, M)$, where $P$ is the set of particle locations and $M$ is the corresponding set of particle masses.
Next, to compute the pressure $\omega_k$ at each particle $k$ as described in equation~\eqref{eq:pressure}, we can use an elementwise subtraction to compute $\rho_k - \rho_0$, a rectified linear unit to compute the $\max$, and finally an elementwise multiplication to multiply by $\lambda_\omega$.
This results in $\Omega$, the set containing the pressure for every particle.

Plugging these values into equation~\eqref{eq:pressureCorrection} is not as straightforward.
It is not obvious how the term $n_{ji}(\omega_i + \omega_j)$ could be represented by $Y$ from the ConvSP equation.
However, by unfolding the terms and distributing the sum we can represent equation~\eqref{eq:pressureCorrection} using ConvSP.

First, note that the vector $n_{ji}$ is simply the difference in position between particles $i$ and $j$ divided by their distance. 
Thus we can replace $n_{ji}$ as follows 
\[ \scalebox{0.8}{${\delta}p^{\omega}_i = \displaystyle\sum_{j \in P-\{i\}} \frac{p_i - p_j}{d_{ij}}(\omega_i + \omega_j)W_\omega(d_{ij}, h)$} \]
where $p_k$ is the location of particle $k$.
For simplicity, let us incorporate the denominator $d_{ij}$ into $W_\omega$ to get it out of the way.
We define $\overline{W}_\omega(d_{ij}, h) = \frac{1}{d_{ij}}W_\omega(d_{ij}, h)$.
% This results in 
% \[ {\delta}p^{\omega}_i = \displaystyle\sum_{j \in P-\{i\}} (p_i - p_j)(\omega_i + \omega_j)\overline{W}_\omega(d_{ij}, h). \]

Next we distribute the terms in the parentheses to get
\[ \scalebox{0.8}{${\delta}p^{\omega}_i = \displaystyle\sum_{j \in P-\{i\}} (p_i\omega_i + p_i\omega_j - p_j\omega_i - p_j\omega_j)\overline{W}_\omega(d_{ij}, h).$}  \]
We can now rearrange the summation and distribute $\overline{W}_\omega$ to yield
\[ \scalebox{0.8}{${\delta}p^{\omega}_i = p_i\omega_i\sum \overline{W}_\omega(d_{ij}, h) + p_i\sum \omega_j\overline{W}_\omega(d_{ij}, h) - \omega_i\sum p_j\overline{W}_\omega(d_{ij}, h) - \sum p_j\omega_j\overline{W}_\omega(d_{ij}, h).$} \]
Here we omitted the summation term $j \in P-\{i\}$ from our notation for clarity.
We can compute this over all $i$ using the ConvSP layer as follows 
\[ \scalebox{0.9}{${\Delta}P^\omega = P*{\Omega}*ConvSP(P, \{1\}) + P*{ConvSP}(P, \Omega) - \Omega*{ConvSP}(P, P) - ConvSP(P, P*{\Omega})$} \]
where $*$ represents elementwise multiplication and $+$ and $-$ are elementwise addition and subtraction respectively.
$\{1\}$ is a set containing all 1s.

\subsection{ConvSDF}

The second layer we add is the ConvSDF layer.
This layer is designed specifically to compute interactions between the particles and static objects in the scene (line~\ref{line:solveCollisions} in figure~\ref{fig:pbf}).
We represent these static objects using signed distance functions (SDFs).
The value $SDF(p)$, where $p$ is a point in space, is defined as the distance from $p$ to the closest point on the object's surface.
If $p$ is inside the object, then $SDF(p)$ is negative.

We define $K$ to be the set of offsets for a given convolutional kernel.
For example, for a $1{\times}3$ kernel in 2D, $K = \{(0,-1), (0,0), (0,1)\}$.
ConvSDF is defined as 
\[ \scalebox{0.8}{$ConvSDF(X) = \Bigg\{\displaystyle\sum_{k \in K} w_k \min_j SDF_j(p_i + k*d) \;\bigg|\; i \in X \Bigg\}$} \]
where $w_k$ is the weight associated with kernel cell $k$, $p_i$ is the location of particle $i$, $SDF_j$ is the $j$th SDF in the scene (one per rigid object), and $d$ is the dilation of the kernel (i.e., how far apart the kernel cells are from each other).
Intuitively, ConvSDF places a convolutional kernel around each particle, evaluates the SDFs at each kernel cell, and then convolves those values with the kernel.
The result is a single value for each particle.

We can use ConvSDF to solve object collisions as follows.
First, we construct $ConvSDF_R$ which uses a size 1 kernel (that is, a convolutional kernel with exactly 1 cell).
We set the weight for the single cell in that kernel to be 1.
With a size 1 kernel and a weight $w_k$ of 1, the summation, the kernel weight $w_k$, and the term $k*d$ fall out of the ConvSDF equation (above).
The result is the SDF value at each particle location, i.e., the distance to the closest surface, where negative values indicate particles that have penetrated inside an object.
We can compute that penetration $R$ of the particles inside objects as 
\[ R = ReLU(-ConvSDF_R(P))  \]
where $ReLU$ is a rectified linear unit.
$R$ now contains the minimum distance each particle would need to move to be placed outside an object, or 0 if the particle is already outside the object.
Next, to determine which direction to ``push'' penetrating particles so they no longer penetrate, we need to find the direction to the surface of the object.
Without loss of generality, we describe how to do this in 3D, but this method is applicable to any dimensionality.
We construct $ConvSDF_X$, which uses a $3{\times}1{\times}1$ kernel, i.e., 3 kernel cells all placed in a line along the X-axis.
We set the kernel cell weights $w_k$ to -1 for the cell towards the negative X-axis, +1 for the cell towards the positive X-axis, and 0 for the center cell.
We construct $ConvSDF_Y$ and $ConvSDF_Z$ similarly for the Y and Z axes.
By convolving each of these 3 layers, we use local differencing in each of the X, Y, and Z dimensions to compute the normal of the surface of the object $n_{SDF}$, i.e., the direction to ``push'' the particle in.
We can then update the particle positions $P'$ as follows
\[ P' = P' +  R*n_{SDF}.  \]
That is, we multiply the distance each particle is penetrating an object ($R$) by the direction to move it in ($n_{SDF}$) and add that to the particle positions.

\subsection{Smooth Particle Networks (SPNets)}

\definecolor{c1}{rgb}{0 0 1}
\definecolor{c2}{rgb}{0.5 0 1}
\definecolor{c3}{rgb}{1 0 1}
\definecolor{c4}{rgb}{1 0.5 0}
\definecolor{c5}{rgb}{0.8 0.8 0}
\definecolor{c6}{rgb}{1 0 0}
\definecolor{c7}{rgb}{0 1 0}
\definecolor{c8}{rgb}{0 0.5 0.2}
\definecolor{c9}{rgb}{0 0.8 0.8}
\newlength{\objectsize}
\setlength{\objectsize}{2.5cm}
\begin{wrapfigure}[30]{R}{8.0cm}
    \centering
    \setlength{\fboxsep}{0pt}
    \setlength{\fboxrule}{1pt}
    \setlength{\unitlength}{1.0cm}
    \begin{subfigure}{8.0cm}
        \vspace{-0.2cm}
        \fbox{\includegraphics[width=\objectsize,height=\objectsize]{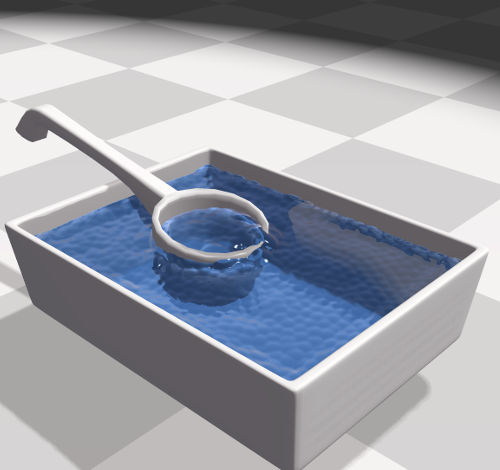}}\hspace{0.1cm}\fbox{\includegraphics[width=\objectsize,height=\objectsize]{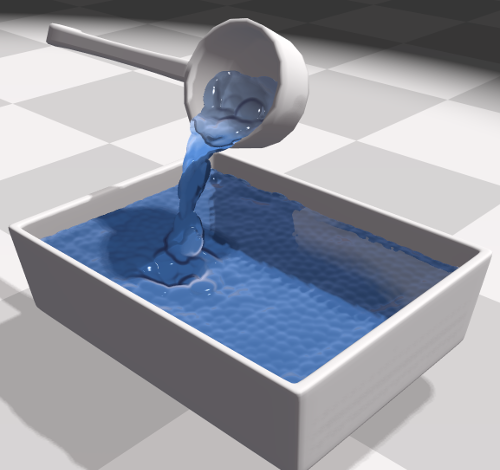}}\hspace{0.1cm}\fbox{\includegraphics[width=\objectsize,height=\objectsize]{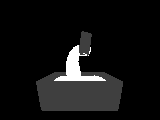}}
        \caption{{\it Ladle} Scene}
        \label{fig:ladle}
    \end{subfigure}
    
    \begin{subfigure}{3.3cm}
        \scalebox{1.1}{
        \begin{picture}(3.0,5.0)
            \put(0.0,2.5){\includegraphics[width=3.0cm]{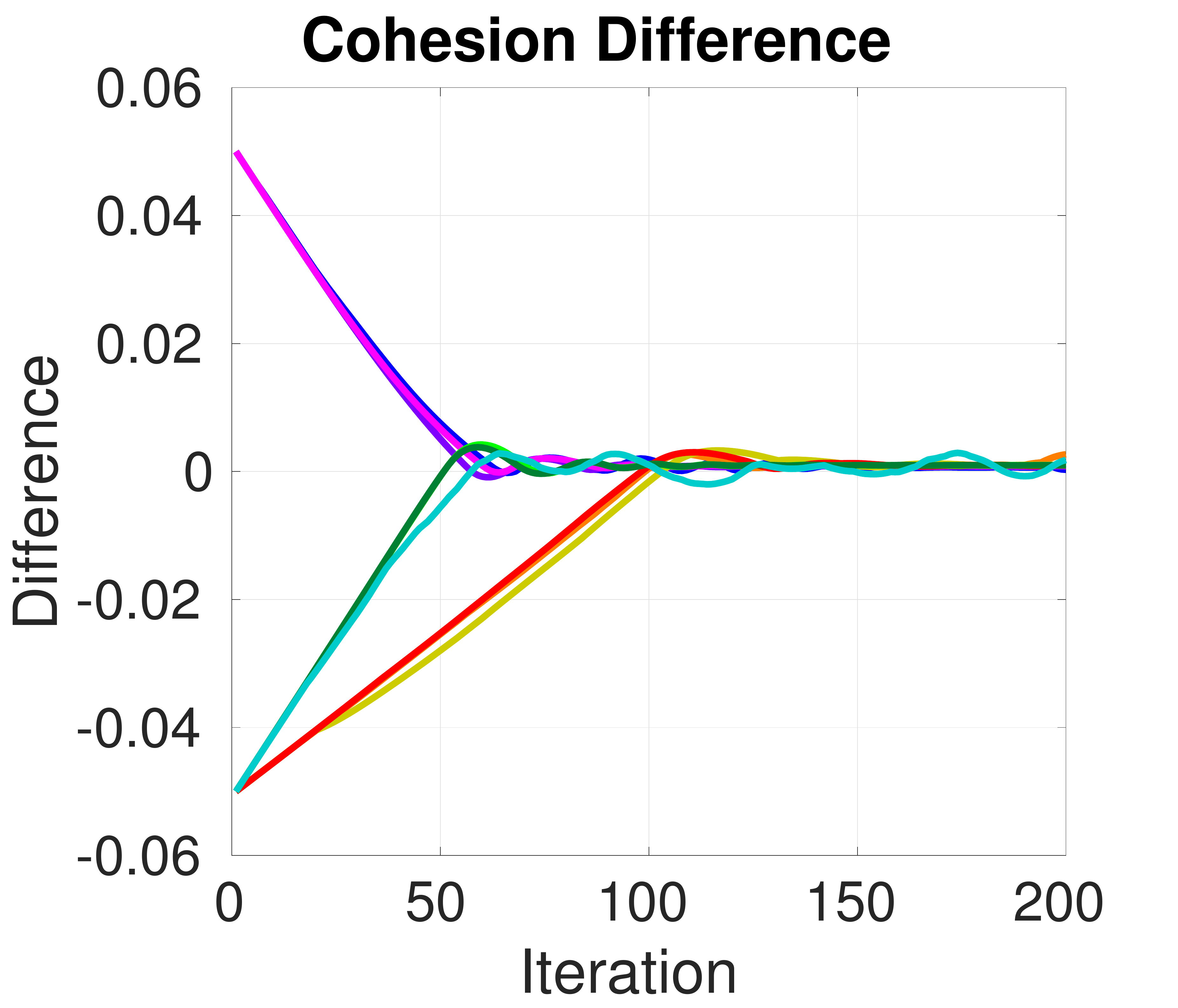}}
            \put(0.15,0){\includegraphics[width=2.82cm]{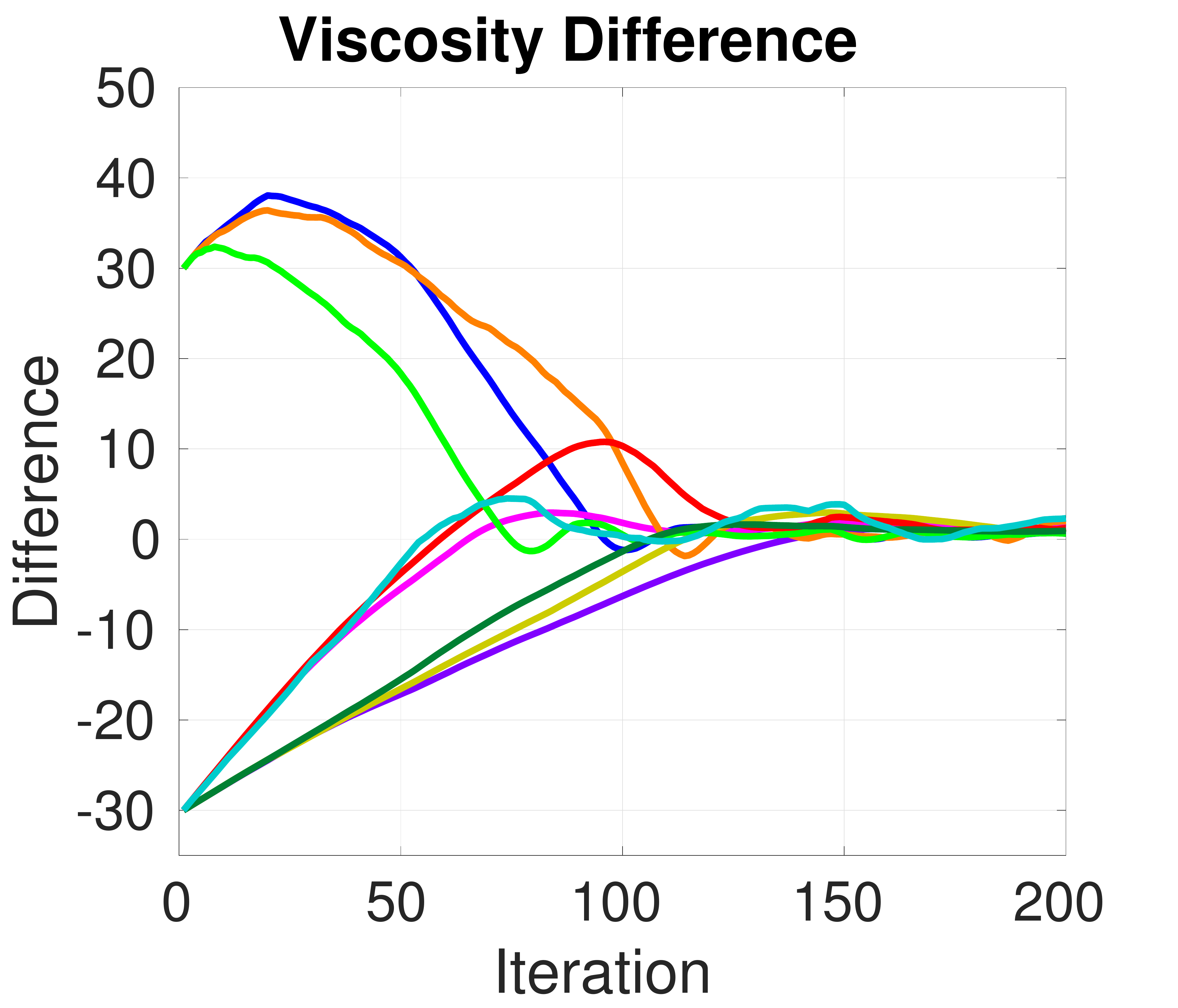}}
        \end{picture}}
        \caption{L1-Loss}
        \label{fig:ex2_l1}
    \end{subfigure}%
    \setlength{\fboxsep}{1pt}%
    \setlength{\fboxrule}{0.5pt}%
    \scalebox{0.7}{\fbox{\parbox{2.0cm}{\centering {\bf\large Legend}\\
        Ground Truth\\
        \hspace{0.8cm} $\lambda_c$, $\lambda_v$\\
        \colorbox{c1}{\parbox[b][0.2cm][s]{0.5cm}{\hspace{0.5cm}}} 0.05, 30\\
        \colorbox{c2}{\parbox[b][0.2cm][s]{0.5cm}{\hspace{0.5cm}}} 0.05, 60\\
        \colorbox{c3}{\parbox[b][0.2cm][s]{0.5cm}{\hspace{0.5cm}}} 0.05, 90\\
        \colorbox{c4}{\parbox[b][0.2cm][s]{0.5cm}{\hspace{0.5cm}}} 0.10, 30\\
        \colorbox{c5}{\parbox[b][0.2cm][s]{0.5cm}{\hspace{0.5cm}}} 0.10, 60\\
        \colorbox{c6}{\parbox[b][0.2cm][s]{0.5cm}{\hspace{0.5cm}}} 0.10, 90\\
        \colorbox{c7}{\parbox[b][0.2cm][s]{0.5cm}{\hspace{0.5cm}}} 0.15, 30\\
        \colorbox{c8}{\parbox[b][0.2cm][s]{0.5cm}{\hspace{0.5cm}}} 0.15, 60\\
        \colorbox{c9}{\parbox[b][0.2cm][s]{0.5cm}{\hspace{0.5cm}}} 0.15, 90}}}%
    \begin{subfigure}{3.3cm}
        \scalebox{1.1}{
        \begin{picture}(3.0,5.0)
            \put(0.0,2.5){\includegraphics[width=3.0cm]{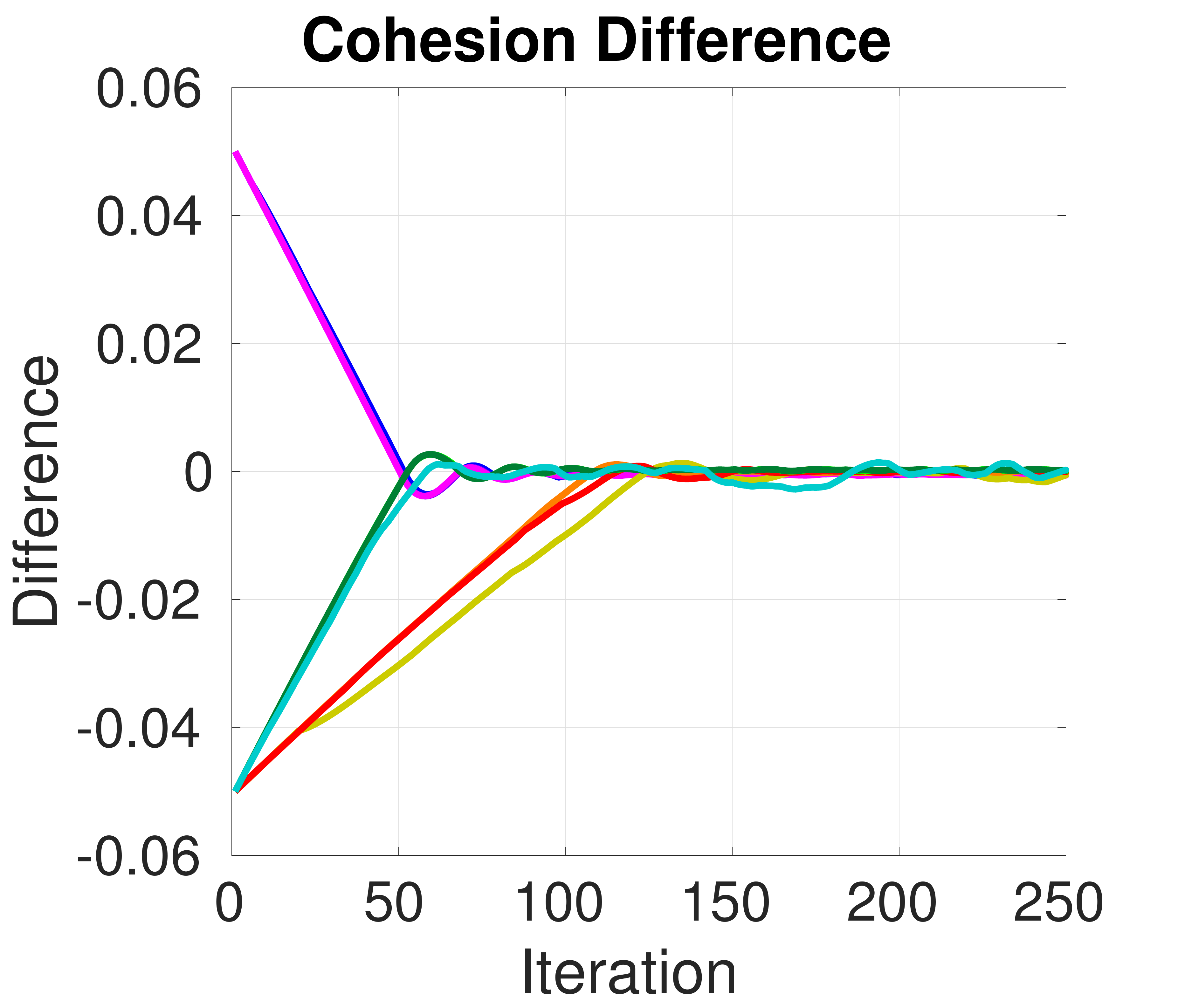}}
            \put(0.15,0){\includegraphics[width=2.82cm]{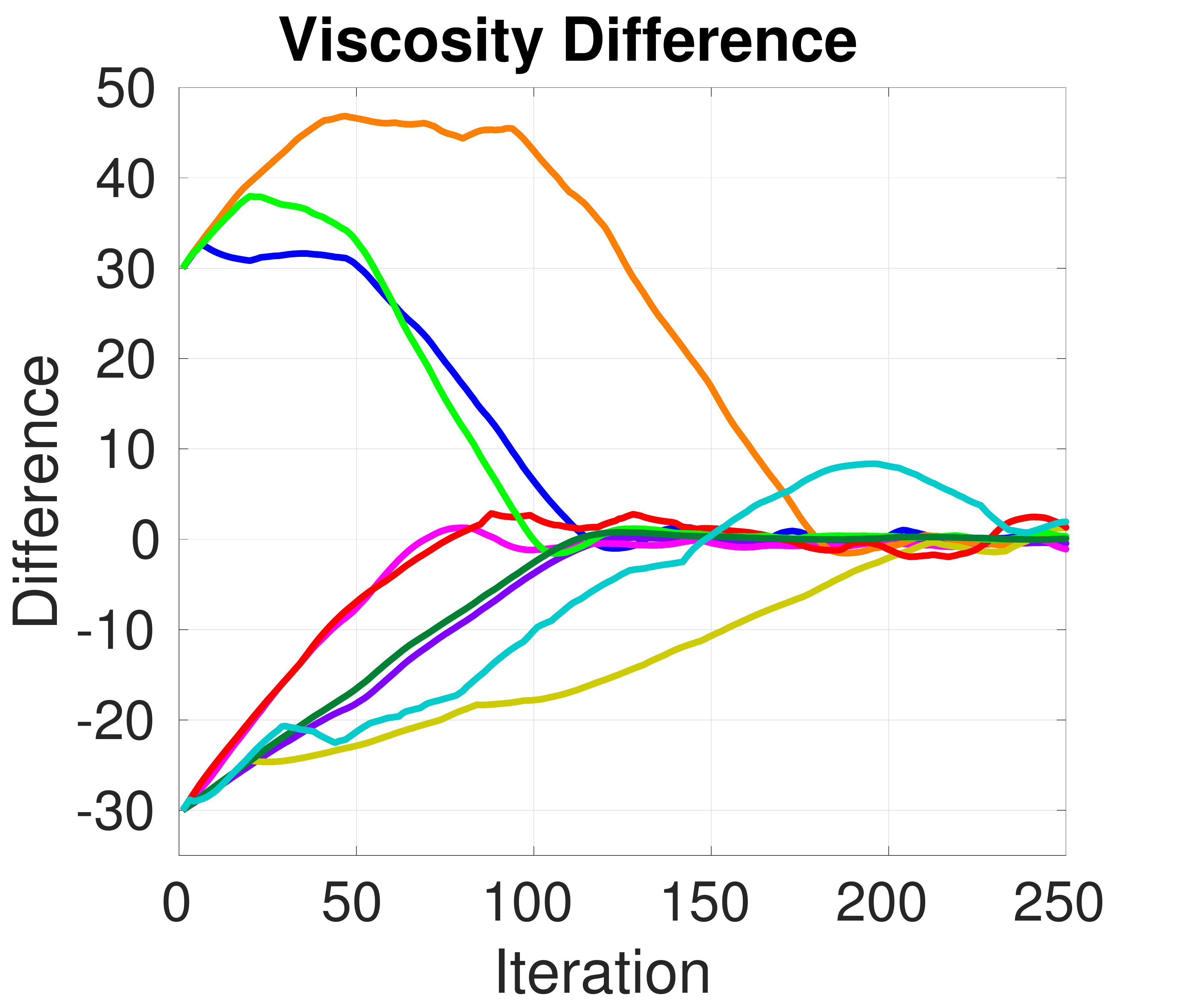}}
        \end{picture}}
        \caption{Projection-Loss}
        \label{fig:ex2_proj}
    \end{subfigure}
    \caption{(Top) The {\it ladle} scene. The left two images are before and after snapshots, the right image shows the particles projected onto a virtual camera image (with the objects shown in dark gray for reference). (Bottom) The difference between the estimated and ground truth fluid parameter values for cohesion $\lambda_c$ and viscosity $\lambda_v$ after each iteration of training for both the L1 loss and the projection loss. The color of the lines indicates the ground truth $\lambda_c$ and $\lambda_v$ values.}
    \label{fig:ex2}
\end{wrapfigure}

Using the ConvSP and ConvSDF layers described in the previous sections and standard network layers, we design SPNets to exactly replicate the PBF algorithm in figure~\ref{fig:pbf}.
That is, at each timestep, the network takes in the current particle positions $P$ and velocities $V$ and computes the fluid dynamics by applying the algorithm line-by-line, resulting in new positions $P'$ and velocities $V'$.
We show an SPNet layout diagram in the appendix.
By repeatedly applying the network to the new positions and velocities at each timestep, we can simulate the flow of liquid over time.
We utilize elementwise layers, rectified linear layers (ReLU), and our two particle layers ConvSP and ConvSDF to compute each line in figure~\ref{fig:pbf}.
Since elementwise and ReLU layers are differentiable, and because we implement analytical gradients for ConvSP and ConvSDF, we can use backpropagation through the whole network to compute the gradients.
Additionally, our layers are implemented with graphics processor support, which means that a forward pass through our network takes approximately $\frac{1}{15}$ of a second for about 9,000 particles running on an Nvidia Titan Xp graphics card.

\section{Evaluation \& Results}
\label{sec:eval}

To demonstrate the utility of SPNets, we evaluated it on three types of tasks, described in the following sections.
First, we show how our model can learn fluid parameters from data.
Next, we show how we can use the analytical gradients of our model to do liquid control.
Finally, we show how we can use SPNets to train a reinforcement learning policy to solve a liquid manipulation task using policy gradients.
Additionally, we also show preliminary results combining SPNets with convolutional neural networks for perception.

\subsection{Learning Fluid Parameters}
\label{sec:ex2}

We evaluate SPNets on the task of learning, or estimating, some of the $\lambda_*$ fluid parameters from data. This experiment illustrates how one can perform system identification on liquids using gradient-based optimization.
Here we frame this system identification problem as a learning problem so that we can apply learning techniques to discover the parameters.
We use a commercial fluid simulator to generate the data and then use backpropagation to estimate the fluid parameters.
We refer the reader to the appendix for more details on this process.

We used the {\it ladle} scene shown in Figure~\ref{fig:ladle} to test our method.
Here, the liquid rests in a rectangular container as a ladle scoops some liquid from the container and then pours it back into the container.
Figures~\ref{fig:ex2_l1} and \ref{fig:ex2_proj} show the difference between the ground truth and estimated values for the cohesion $\lambda_c$ and viscosity $\lambda_v$ parameters when using the model to estimate the parameters on each of the 9 sequences we generated, for both the L1-loss, which assumes full knowledge of the particle state, and the projection loss, which assumes the system has access to only a 2D projection of the particle state.
In all 9 cases and for both losses, the network converges to the ground truth parameter values after only a couple hundred iterations.
While the L1 loss tended to converge slightly faster (which is to be expected with a more dense loss signal), the projection loss was equally able to converge to the correct parameter values, indicating that the gradients computed by our model through the camera projection are able to correctly capture the changes in the liquid due to its parameters.
Note that for the projection loss the camera position is important to provide the silhouette information necessary to infer the liquid parameters.

\subsection{Liquid Control}
\label{sec:control}

\setlength{\objectsize}{2.5cm}
\begin{wrapfigure}[16]{R}{7.9cm}
    \vspace{-1.5cm}
    \centering
    \setlength{\fboxsep}{0pt}
    \setlength{\fboxrule}{1pt}
    \setlength{\unitlength}{1.0cm}
    \begin{subfigure}{\objectsize}
        \fbox{\includegraphics[width=\objectsize,height=\objectsize]{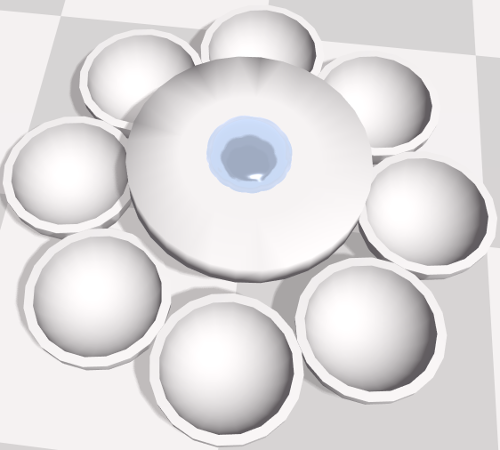}}
        \fbox{\includegraphics[width=\objectsize,height=\objectsize]{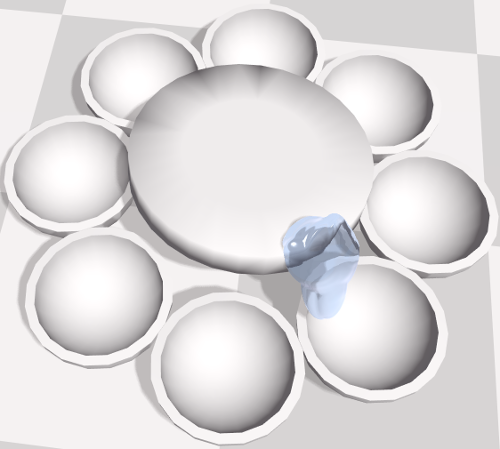}}
        \caption{{\it Plate}}
        \label{fig:plate}
    \end{subfigure}\hspace{0.2cm}%
    \begin{subfigure}{\objectsize}
        \fbox{\includegraphics[width=\objectsize,height=\objectsize]{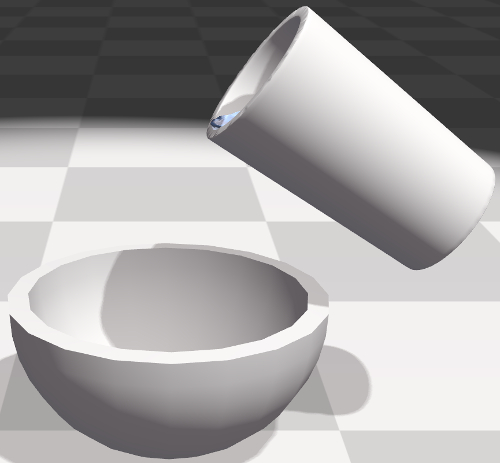}}
        \fbox{\includegraphics[width=\objectsize,height=\objectsize]{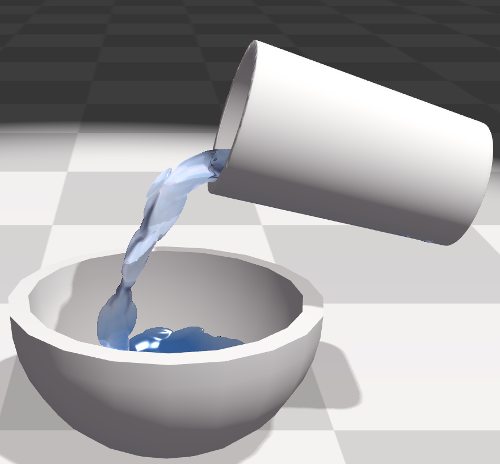}}
        \caption{{\it Pouring}}
        \label{fig:pouring}
    \end{subfigure}\hspace{0.2cm}%
    \begin{subfigure}{\objectsize}
        \fbox{\includegraphics[width=\objectsize,height=\objectsize]{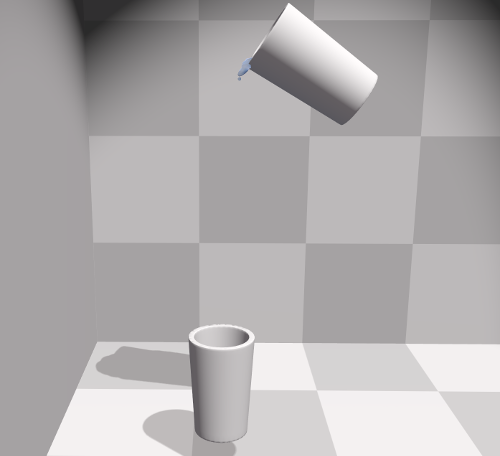}}
        \fbox{\includegraphics[width=\objectsize,height=\objectsize]{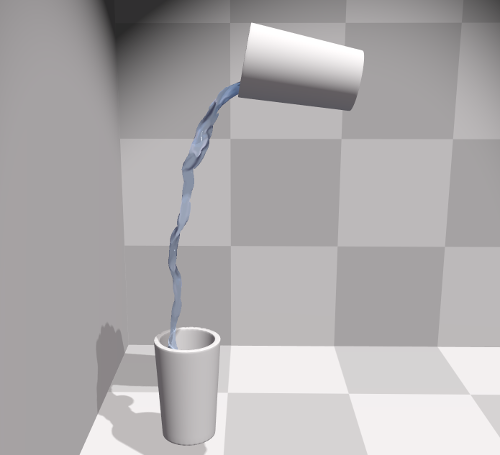}}
        \caption{{\it Catching}}
        \label{fig:catching}
    \end{subfigure}
    \caption{The control scenes used in the evaluations in this paper. The top row shows the initial scene; the bottom row shows the scene after several seconds.}
    \label{fig:control_scenes}
\end{wrapfigure}

To test the efficacy of the gradients produced by our models, we evaluate SPNets on 3 liquid control problems.
The goal in each is to find the set of controls $\mathcal{U} = \{u_t\}$ that minimize the cost
\begin{equation}
L = \displaystyle\sum_t l(P_t, V_t, u_t) \label{eq:mpc_loss}
\end{equation}
where $l$ is the cost function, $P_t$ is the set of particle positions at time $t$, and $V_t$ is the set of particle velocities at time $t$.
To optimize the controls $\mathcal{U}$, we utilize Model Predictive Control (MPC)~\cite{camacho2013}.
MPC optimizes the controls for a short, finite time horizon, and then re-optimizes at every timestep.
We evaluated our model on 3 scenes: the {\it plate} scene, the {\it pouring} scene, and the {\it catching} scene.
We refer the reader to the appendix for details on how MPC is used to optimize the controls for each scene.

\begin{figure}
  \centering
  \begin{subfigure}{3.0cm}
    \setlength{\fboxsep}{0pt}
    \setlength{\fboxrule}{1pt}
    \setlength{\unitlength}{1.0cm}
    \scalebox{0.6}{
        \begin{picture}(5.0,5.0)
            \put(0.0,0.0){\fbox{\includegraphics[width=5.0cm]{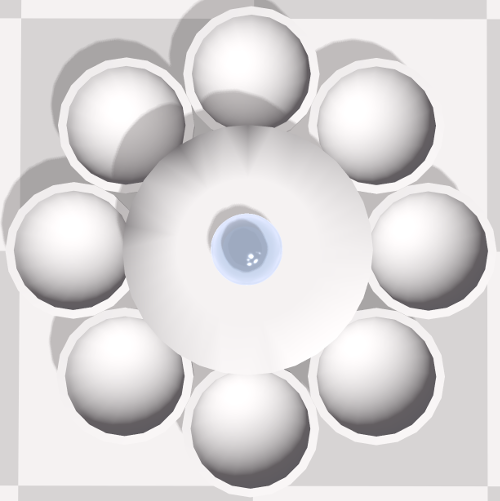}}}
            \put(3.9, 2.42){{\color{red}\bf 88.9\%}}
            \put(3.4, 3.7){{\color{red}\bf 100\%}}
            \put(2.15, 4.2){{\color{red}\bf 100\%}}
            \put(0.9, 3.7){{\color{red}\bf 100\%}}
            \put(0.3, 2.42){{\color{red}\bf 100\%}}
            \put(0.9, 1.14){{\color{red}\bf 100\%}}
            \put(2.15, 0.64){{\color{red}\bf 100\%}}
            \put(3.4, 1.14){{\color{red}\bf 100\%}}
        \end{picture}}
        \caption{{\it Plate} Scene}
        \label{fig:plate_results}
  \end{subfigure}\hspace{0.2cm}%
  \begin{subfigure}{4.0cm}
        \includegraphics[width=4.0cm]{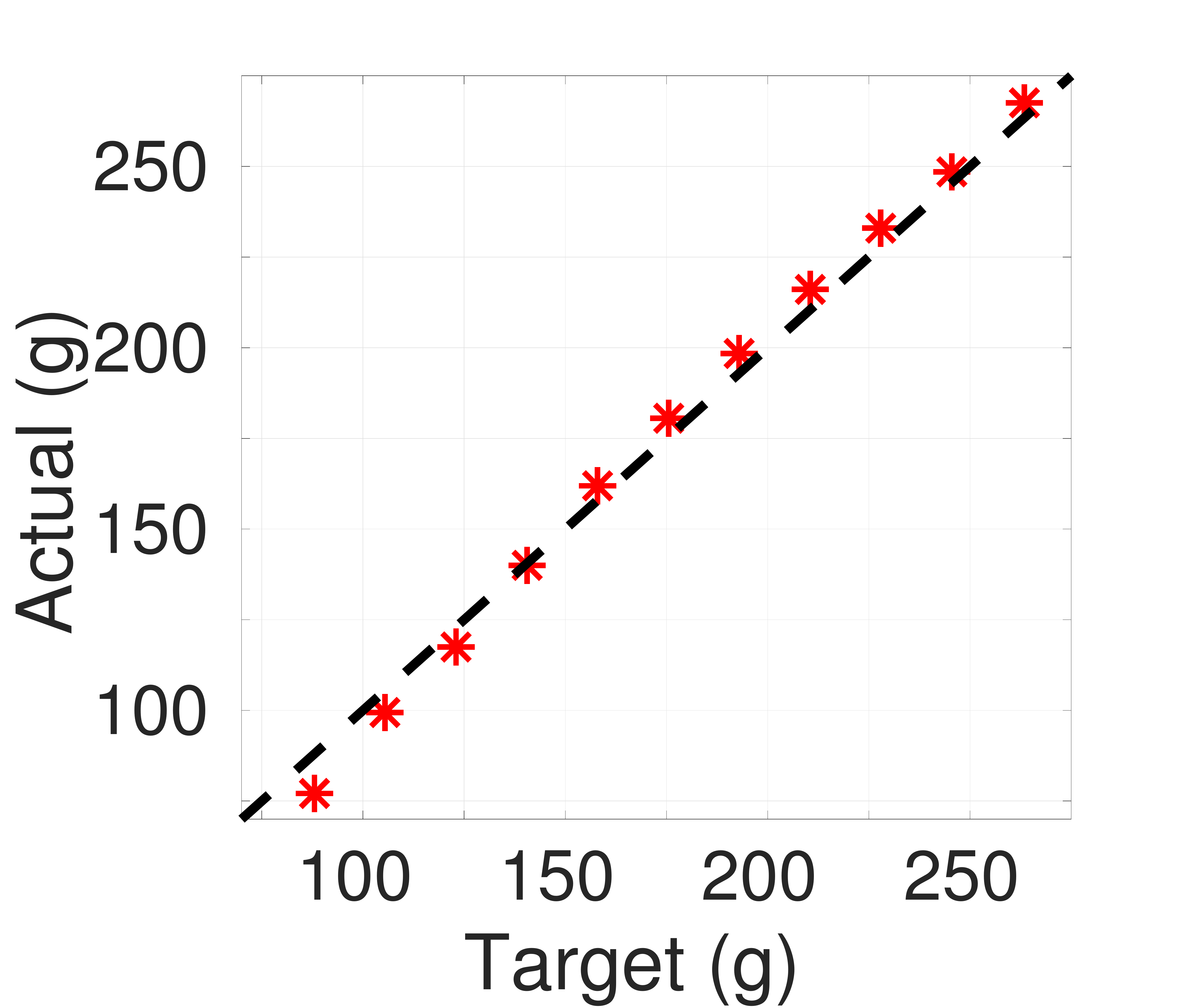}
        \caption{{\it Pouring} Scene}
        \label{fig:pour_results}
  \end{subfigure}%
  \begin{subfigure}{5.0cm}
        \begin{tabular}{ r || c | c | c }
             Initial & & Policy & Policy \\
             Movement & MPC & Train & Test\\\hline\hline
             Right & 97.9\% & 98.3\% & 99.2\% \\
             Left & 99.7\% & 99.5\% & 93.8\% \\\hline
             Both & 98.8\% & 98.9\% & 96.5\%
        \end{tabular}
        \caption{{\it Catching} Scene}
        \label{fig:catching_results}
  \end{subfigure}
  \caption{Results from the liquid control task. From left to right: The {\it plate} scene. The numbers in each bowl indicate the percent of particles that were successfully placed in that bowl when that bowl was the target. The {\it pouring} scene. The x axis is the targeted pour amount and the y axis is the amount of liquid that was actually poured where the red marks indicate each of the 11 pours. The {\it catching} scene. Shown is the percent of liquid caught by the target cup where the rows indicate the initial direction of movement of the source.}
    \label{fig:combined_results}
    \vspace{-0.6cm}
\end{figure}
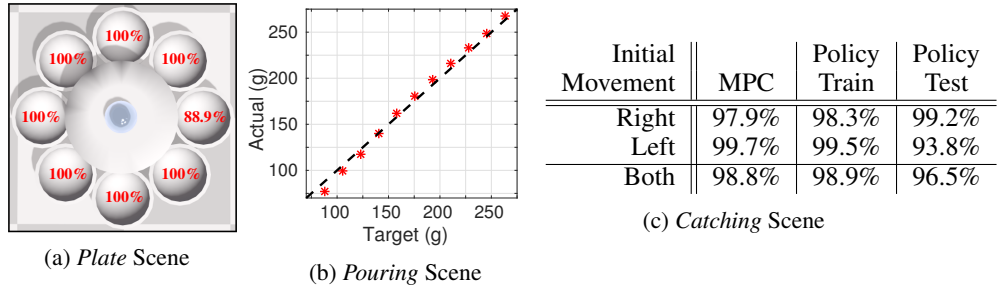

\textbf{The {\it Plate} Scene:} 
Figure~\ref{fig:plate} shows the {\it plate} scene.
It consists of a plate surrounded by 8 bowls.
A small amount of liquid is placed on the center of the plate, and the plate must be tilted such that the liquid falls into a given target bowl.
Figure~\ref{fig:plate_results} shows the results of each of the evaluations on the {\it plate} scene.
In every case, the optimization produced a trajectory where the plate would ``dip'' in the direction of the target bowl, wait until the liquid had gained sufficient momentum, and then return upright, which allowed the liquid to travel further off the edge of the plate.
Note that simply ``dipping'' the plate would result in the liquid falling short of the bowl.
For all bowls except one, this resulted in 100\% of the liquid being placed into the correct bowl.
For the one bowl, when it was set as the target, all but a small number of the liquid particles were placed in the bowl.
Those particles landed on the lip of the bowl, eventually rolling off onto the ground.
Nonetheless, it is clear that our method is able to effectively solve this task in the vast majority of cases.

%\subsubsection{The {\it Pouring} Scene}

\textbf{The {\it Pouring} Scene:}
We also evaluated our method on the {\it pouring} scene, shown in Figure~\ref{fig:pouring}.
The goal of this task is to pour liquid from the cup into the bowl.
In all cases we evaluated, all liquid either remained in the cup or was poured into the bowl; no liquid was spilled on the ground.
For that reason, in figure~\ref{fig:pour_results} we show how close each evaluation was to the given desired pour amount.
In every case, the amount poured was within 11g of the desired amount, and the average difference across all 11 runs between actual and desired was 5g.
Note that the initial rotation of the cup happens implicitly; our loss function only specifies a desired target for the liquid, not any explicit motion.
This shows that physical reasoning about fluids using our model enables solving fine-grained reasoning tasks like this.

\textbf{The {\it Catching} Scene:} The final scene we evaluated on was the {\it catching} scene, shown in Figure~\ref{fig:catching}.
The scene consisted of two cups, a source cup in the air filled with liquid and a target cup on the ground.
The source cup moved arbitrarily while dumping the liquid in a stream.
The goal of this scene is to shift the target cup along the ground to catch the stream of liquid and prevent it from hitting the ground.
The first column of the table in Figure~\ref{fig:catching_results} shows the percentage of liquid caught by the cup averaged across our evaluations.
In all cases, the vast majority of the liquid was caught, with only a small amount dropped due largely to the time it took the target cup to initially move under the stream.
It is clear from these results and the liquid control results on the previous two scenes that our model can enable fine-grained reasoning about fluid dynamics.

\subsection{Learning a Liquid Control Policy via Reinforcement Learning}

Finally, we evaluate our model on the task of learning a policy in a reinforcement learning setting.
That is, the control $u_t$ at timestep $t$ is computed as a function of the state of the environment, rather than optimized directly as in the previous section.
Here the goal of the robot is to optimize the parameters of the policy.
We refer the reader to the appendix for details on the policy training procedure.
We test our methodology on the {\it catching} scene.
The middle column of the table in figure~\ref{fig:catching} shows the percent of liquid caught by the target cup when using the policy to generate the controls for the training sequences.
In all cases, the vast majority of the liquid was caught by the target cup.
To test the generalization ability of the policy, we modified the sequences as follows.
For all the training sequences, the source cup rotated counter-clockwise (CCW) when pouring.
To test the policy, we had the source cup follow the same movement trajectory, but rotated clockwise (CW) instead, i.e., in training the liquid always poured from the left side of the source, but for testing it poured out of the right side.
The percent of liquid caught by the target cup when using the policy for the CW case is shown in the third column of the table in figure~\ref{fig:catching}.
Once again the policy is able to catch the vast majority of the liquid.
The main point of failure is when the source cup initially moves to the left.
In this case, as the source cup rotates, the liquid initially only appears in the upper-left of the image.
It's not until the liquid has traveled downward several centimeters that the policy begins to move the target cup under the stream, causing it to fail to catch the initial liquid.
This behavior makes sense, given that during training the policy only ever encountered the source cup rotating CCW, resulting in liquid never appearing in the upper-left of the image.
Nonetheless, these results show that, at least in this case, our method enables us to train robust policies for solving liquid control tasks.

\subsection{Combining SPNets with Perception}

\setlength{\objectsize}{2.5cm}
\begin{wrapfigure}[25]{R}{7.9cm}
    % \vspace{-0.5cm}
    \centering
    \setlength{\fboxsep}{0pt}
    \setlength{\fboxrule}{1pt}
    \setlength{\unitlength}{1.0cm}
    \begin{subfigure}{\objectsize}
        \fbox{\includegraphics[width=\objectsize,height=\objectsize]{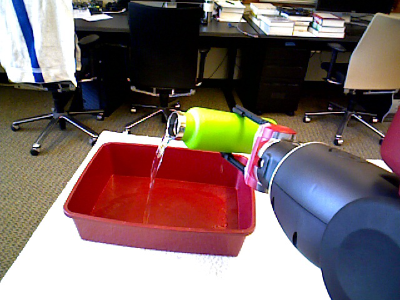}}
        \fbox{\includegraphics[width=\objectsize,height=\objectsize]{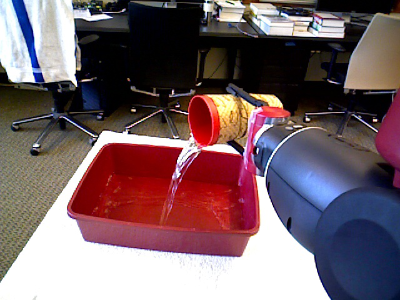}}
        \caption{RGB}
        \label{fig:plate}
    \end{subfigure}\hspace{0.2cm}%
    \begin{subfigure}{\objectsize}
        \fbox{\includegraphics[width=\objectsize,height=\objectsize]{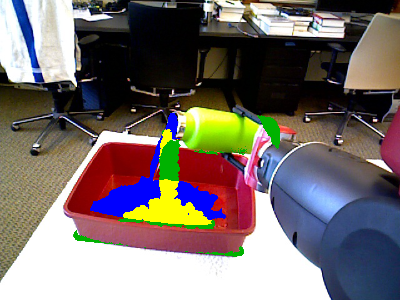}}
        \fbox{\includegraphics[width=\objectsize,height=\objectsize]{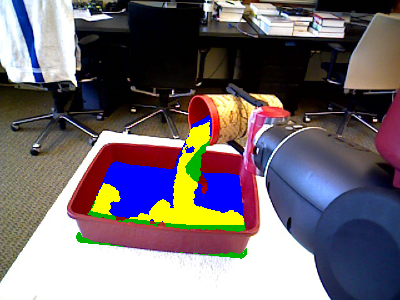}}
        \caption{SPNets}
        \label{fig:pouring}
    \end{subfigure}\hspace{0.2cm}%
    \begin{subfigure}{\objectsize}
        \fbox{\includegraphics[width=\objectsize,height=\objectsize]{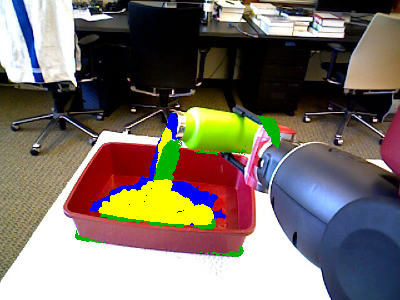}}
        \fbox{\includegraphics[width=\objectsize,height=\objectsize]{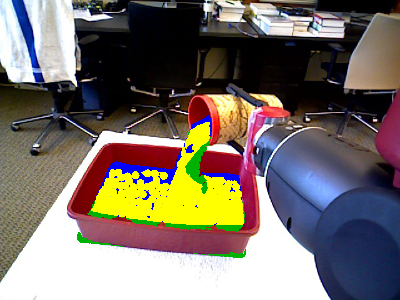}}
        \caption{{\scriptsize SPNets+Perception}}
        \label{fig:catching}
    \end{subfigure}
    \caption{Results when combining SPNets with perception. The images in the top and bottom rows show 2 example frames. From left-to-right: the RGB image (for reference), the RGB image with SPNets overlayed (not using perception), and the RGB image with SPNets with perception overlayed. In the overlays, the blue color indicates ground truth liquid pixels, green indicates the liquid particles, and yellow indicates where they overlap.}
    \label{fig:spnets_perception}
\end{wrapfigure}

While the focus in this paper has been on introducing SPNets as a platform for differentiable fluid dynamics, we also wish to show an example of how it can be combined with real perception.
So in this section we show some initial results on a liquid state tracking task.
That is, we use a real robot and have it interact with liquid.
During the interaction, the robot observes the liquid via its camera.
It is then tasked with reconstructing the full 3D state of the liquid at each point in time from this observation.
To make this task feasible, we assume that the initial state of the liquid is known and that the robot has 3D models of and can track the 3D pose of the rigid objects in the scene.
The robot then uses SPNets to compute the updated liquid position at each point in time, and uses the observation to correct the liquid state when the dynamics of the simulated and real liquid begin to diverge.
This is the same task we looked at in~\citep{schenckc2017a}.
We describe the details of this method in the appendix.

We evaluated the robot on 12 pouring sequences.
Figure~\ref{fig:spnets_perception} shows 2 example frames from 2 different sequences and the result of both SPNets and SPNets with perception.
The yellow pixels indicate where the model and ground truth agree; the blue and green where they disagree.
From this it is apparent that SPNets with perception is significantly better at matching the real liquid state than SPNets without perception.
We evaluate the intersection-over-union (IOU) across all frames of the 12 pouring sequences.
SPNets alone (without perception) achieved an IOU of 36.1\%.
SPNets with perception achieved an IOU of 56.8\%.
These results clearly show that perception is capable of greatly improving performance even when there is significant model mismatch.
Here we can see that SPNets with perception increased performance by 20\%, and from the images in Figure~\ref{fig:spnets_perception} it is clear that this increase in performance is significant.
This shows that our framework can be very useful for combining real perceptual input with fluid dynamics.

\vspace{-0.3cm}
\section{Conclusion \& Future Work}
\label{sec:conclusion}
\vspace{-0.3cm}

In this paper we presented SPNets, a method for computing differentiable fluid dynamics and their interactions with rigid objects inside a deep network.
To do this, we developed the ConvSP and ConvSDF layers, which allow the model to compute particle-particle and particle-rigid object interactions.
We then showed how these layers can be combined with standard neural network layers to compute fluid dynamics.
Our evaluation in Section~\ref{sec:eval} showed how a fully differentiable fluid model can be used to 1) learn, or identify, fluid parameters from data, 2) control liquids to accomplish a task, 3) learn a policy to control liquids, and 4) be used in combination with perception to track liquids.
This is the power of model-based methods: they are widely applicable to a variety of tasks.
Combining this with the adaptability of deep networks, we can enable robots to robustly reason about and manipulate liquids.
Importantly, SPNets make it possible to specify liquid identification and control tasks in terms of the \emph{desired state of the liquid}; the resulting controls follow from the physical interaction between the liquid and the controllable objects.  This is in contrast to prior approaches to pouring liquids, for instance, where the relationships between controls and liquid states have to be specified via manually designed functions.

We believe that by combining model-based methods with deep networks for liquids, SPNets provides a powerful new tool to the roboticist's toolbox for enabling robots to handle liquids. 
A possible next step for future work is to add a set of parameters to SPNets to facilitate learning a residual model between the analytical fluid model and real observed fluids, or even to learn the dynamics of different types of substances such as sand or flour.
SPNets can also be used to perform more complex manipulation tasks, such as mixing multiple liquid ingredients in a bowl, online identification and prediction of liquid behavior,
%constructing pipes to direct water to desired positions, 
or using spoons to move liquids, fluids, or granular media between containers.

{\fontsize{9.9}{9.9pt}\selectfont
\bibliography{references}  % .bib
}
\newpage

\appendix
\input{appendix_inner.tex}

\end{document}

%% file: appendix_inner.tex
\section{Position Based Fluids Continued}

In section~3 of the main paper we gave a brief overview of the Position Based Fluids (PBF) algorithm.
The key steps of the PBF algorithm are moving the particles (lines~2--3 in figure~1 in the main paper), iteratively solving the constrains imposed by the incompressibility of the fluid (lines~4--10), and updating the velocities (lines~11--12).
Solving the constraints entails iteratively moving each particle to better satisfy each constraint until the constraints are satisfied.
In PBF, this is approximated by repeating the inner loop (lines~4--10) a fixed number of times.

In the main paper we described some of the details of computing the various constraint solutions.
Here we describe computing the solutions to the \textproc{SolveCohesion} (line~6), \textproc{SolveSurfaceTension} (line~7), and \textproc{ApplyViscosity} (line~12) functions.

The cohesion correction ${\delta}p^c_i$ for each particle $i$ is computed as
\[ {\delta}p^c_i = \displaystyle\sum_{j \in P-\{i\}} \lambda_c n_{ij} W_c(d_{ij}, h) \]
where $\lambda_c$ is the cohesion constant and $W_c$ is a kernel function.
For $W_c$ we use
\[ W_c(d, h) = \frac{-(1 - d_0)}{d_0^2} \left( \frac{d}{h} \right)^3 + \frac{d_0^2 + d_0 + 1}{d_0^2} \left( \frac{d}{h} \right)^2 - 1 \]
where $d_0$ is the fluid rest distance as a fraction of $h$.
For this paper we fix $d_0$ to $0.5$.

The surface tension correction ${\delta}p^s_i$ for each particle $i$ is computed using the following 2 equations 
\[{\delta}p^c_i = \displaystyle\sum_{j \in P-\{i\}} \frac{\lambda_s}{\rho_0}(n_j - n_i)I(d_{ij} \leq h)  \]
where $\lambda_s$ is the surface tension constant, $n_k$ is the normal of the fluid surface at particle $k$, and $I$ is the indicator function.
The normal $n_k$ is computed as 
\[ n_k = \displaystyle\sum_{j \in P-\{i\}} n_{ij} W_c(d_{ij}, h)  \]
where $W_c$ is the same kernel function used for the cohesion constraint.

Finally, the viscosity update ${\delta}v_i$ for each particle $i$ computed by \textproc{ApplyViscosity} is
\[{\delta}v_i = \displaystyle\sum_{j \in P-\{i\}} \frac{\lambda_v}{\rho_0} (v_j - v_i) W_\rho(d_{ij}, h)  \]
where $\lambda_v$ is the viscosity constant, $v_k$ is the velocity of particle $k$, and $W_\rho$ is the same kernel function used to compute the density. 

\section{SPNet Diagram}

\begin{figure}
  \centering
  \begin{subfigure}{8.0cm}
      \includegraphics[width=8.0cm]{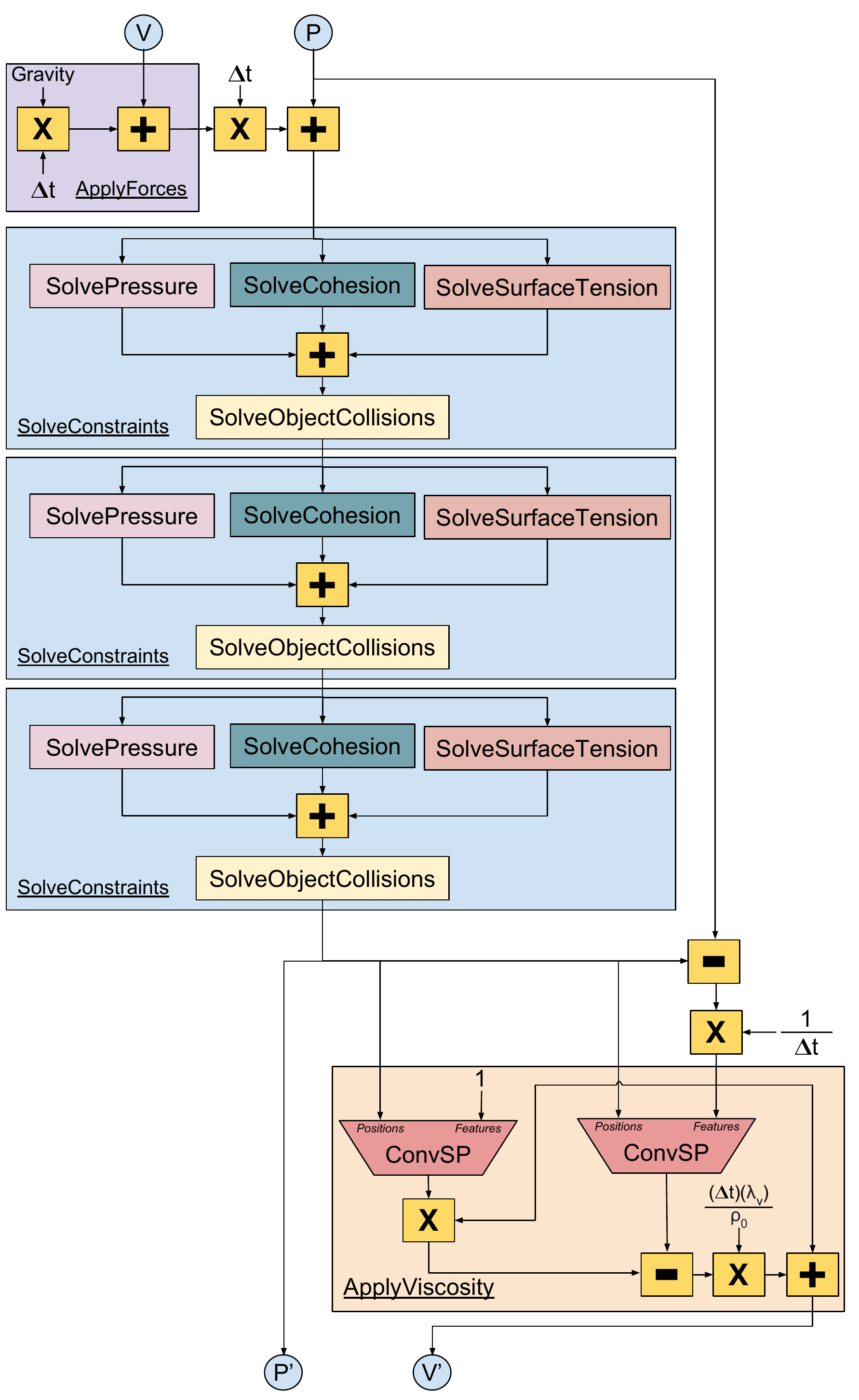}
      \caption{SPNet}
      \label{fig:spnet}
  \end{subfigure}%
  \begin{subfigure}{5.0cm}
      \setlength{\fboxsep}{0pt}
      \setlength{\fboxrule}{1pt}
      \fbox{\includegraphics[width=5.0cm]{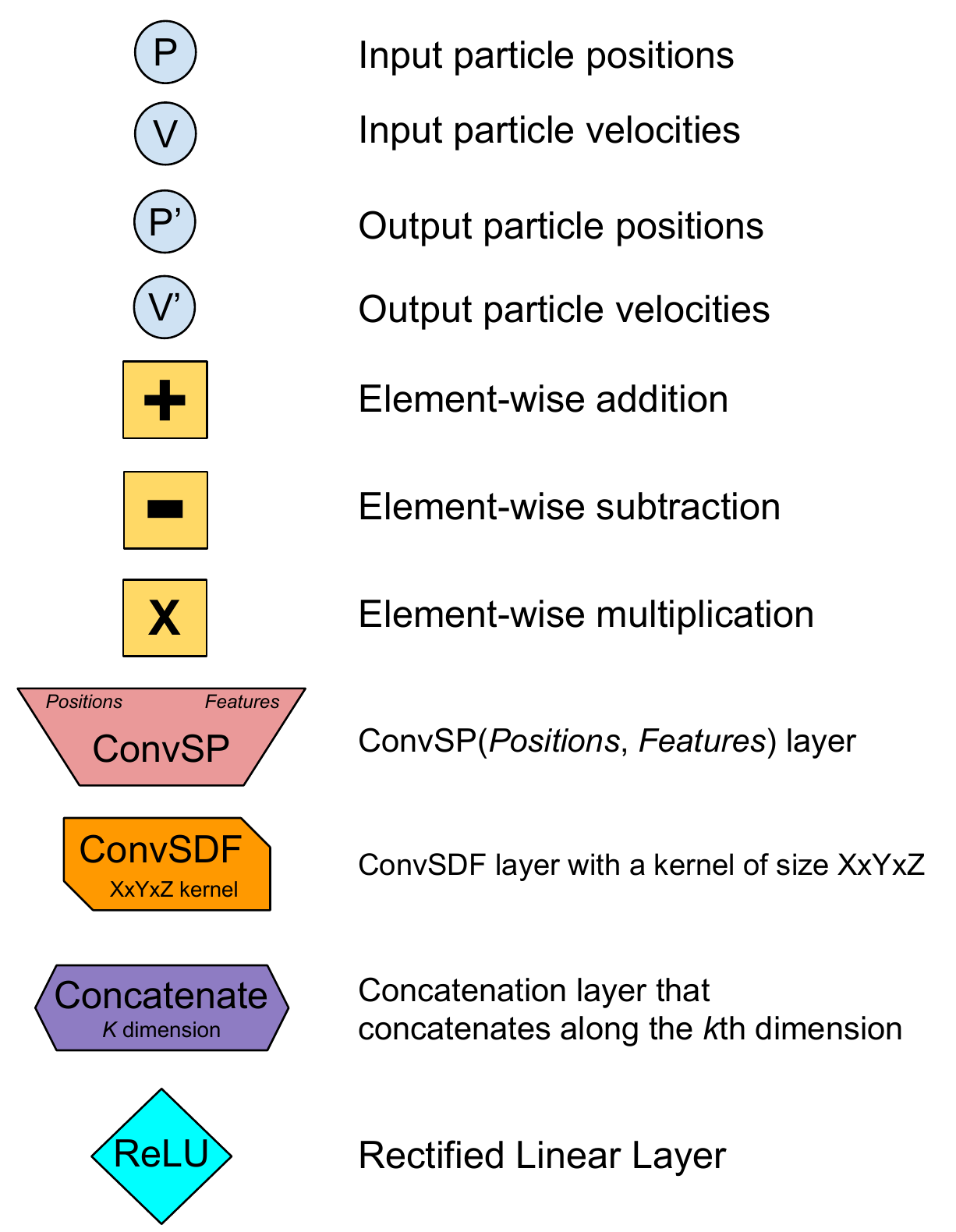}}
      \caption{Legend}
      \label{fig:legend}
      \vspace{6.0cm}
  \end{subfigure}
  
  \vspace{1.0cm}
  
  \begin{subfigure}{7.0cm}
      \includegraphics[width=7.0cm]{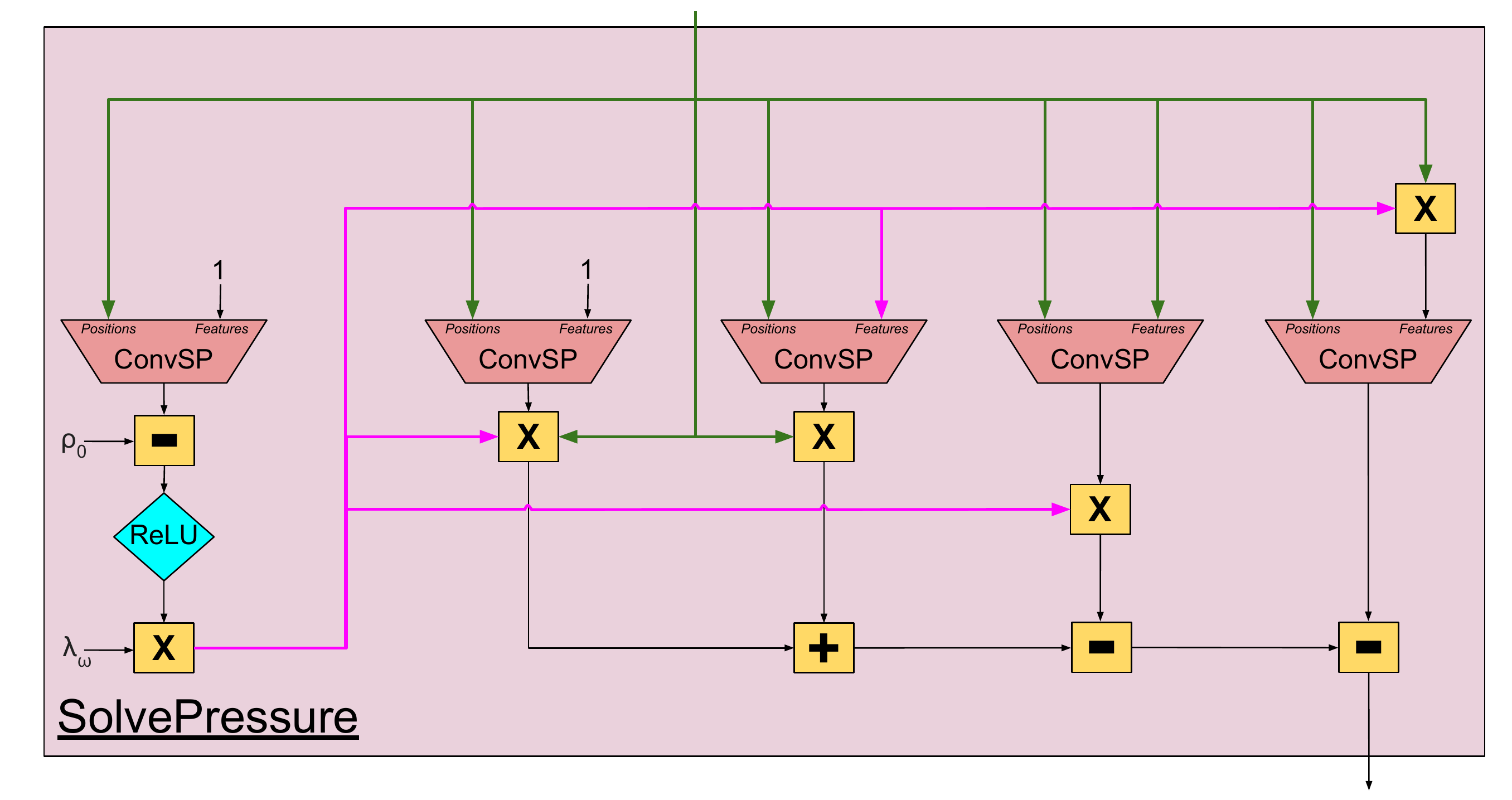}
      \caption{SolvePressure}
      \label{fig:solvepressure}
  \end{subfigure}%
  \begin{subfigure}{7.0cm}
      \includegraphics[width=7.0cm]{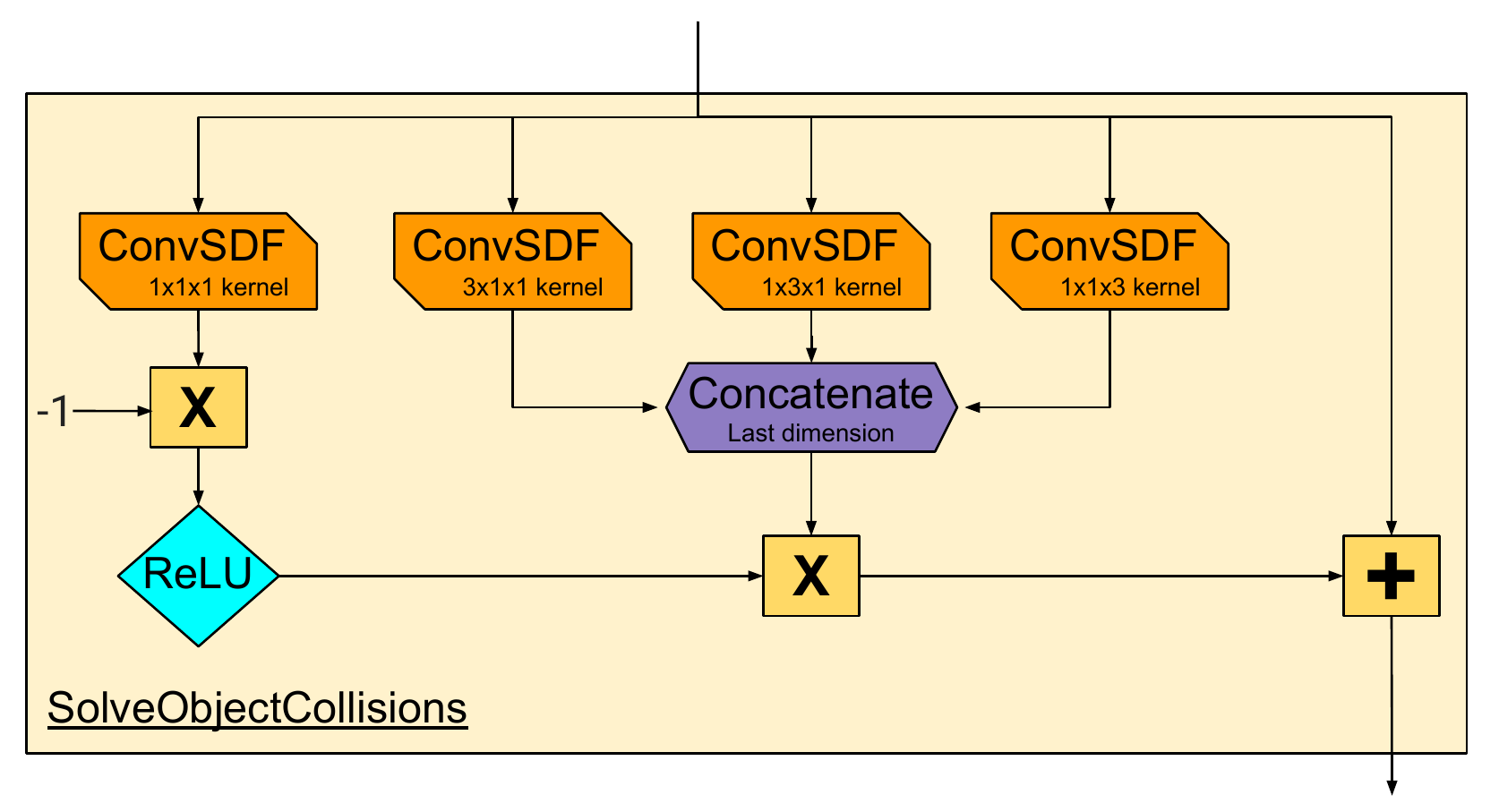}
      \caption{SolveObjectCollisions}
      \label{fig:solveobjectcollisions}
  \end{subfigure}
  \caption{The layout for SPNet. The upper-left shows the overall layout of the network. The functions \textproc{SolvePressure}, \textproc{SolveCohesion}, \textproc{SolveSurfaceTension}, and \textproc{SolveObjectCollisions} are collapsed for readability. The lower-right shows the expansion of the \textproc{SolveObjectCollisions} function, with the line in the top of the box being the input to the \textproc{SolveObjectCollisions} in the upper-left diagram and the line out of the bottom of the box being the line out of the box. The lower-left shows the expansion of the \textproc{SolvePressure} function. For clarity, the input line (which represents the particle positions) is colored green and the line representing the particle pressures is colored pink.}
  \label{fig:net_diagram}
\end{figure}

As described in section~3, Smooth Particle Networks (SPNets) implements the Position Based Fluids (PBF) alrogithm, which is shown in the main paper in figure~1.
Figure~\ref{fig:net_diagram} shows the layout of SPNets as a network diagram.
The network takes as input the current particle positions $P$ and velocities $V$, and computes the fluid dynamics for a single timestep resulting in the new positions $P'$ and velocities $V'$.
For clarity, the fhe functions \textproc{SolvePressure}, \textproc{SolveCohesion}, \textproc{SolveSurfaceTension}, and \textproc{SolveObjectCollisions} are collapsed into individual boxes in figure~\ref{fig:spnet}.
The full layout for \textproc{SolvePressure} and \textproc{SolveObjectCollisions} are shown in figures \ref{fig:solvepressure} and \ref{fig:solveobjectcollisions} respectively.

The first operation the network performs is to apply the external forces to the particles, line~2 of the PBF algorithm (shown in figure~1 in the main paper) and the lavender box in the upper-left of figure~\ref{fig:spnet} here.
Next the network updates the particle positions according to their velocities, line~3 of PBF and the element-wise multiplication and addition immediately to the right of \textproc{ApplyForces}.
After this, the network iteratively solves the fluid constraints (lines 5--9), shown by the \textproc{SolveConstraints} boxes in figure~\ref{fig:spnet}.
Here we show 3 constraint solve iterations, however in principle the network could have any number.
Each constraint solve partially updates the particle positions to better satisfy the given constraints.

We consider 3 constraints in this paper: pressure (line~5), cohesion (line~6), and surface tension (line~7).
Each is shown as an individual box in figure~\ref{fig:spnet}.
Figure~\ref{fig:solvepressure} shows the full network layout for the pressure constraint.
This exactly computes the solutions to equations 1--3 from the main paper as derived in section~4.1.
Note the column under the leftmost ConvSP layer in figure~\ref{fig:spnet}; it computes the pressure set $\Omega$.
This is then used to compute the result of the other 4 ConvSP layers.
The final step of each constraint solve iteration is to solve the object collisions.
The expansion of this box is shown in figure~\ref{fig:solveobjectcollisions}.
The ConvSDF layer on the left computes the particle penetration $R$ into the SDFs, and the 3 on the right compute the normal $n_{SDF}$ of the SDFs.
Note that in this diagram we show the layout for particles in 3D (there are 3 ConvSDF layers on the right of figure~\ref{fig:solveobjectcollisions}, one to compute the normal direction in each dimension), however this can applied to particles in any dimensionality.

After finishing the constraint solve iterations, the network computes the adjusted particle velocities based on how the positions were adjusted (line~11 of the PBF algorithm), shown in figure~\ref{fig:spnet} as the element-wise subtraction and multiplication above the \textproc{ApplyViscosity} box.
Finally, the network computes the viscosity, shown in the tan box in the bottom-right of figure~\ref{fig:spnet}.
Viscosity only affects the particles velocities, so the output positions of the particles are the same as computed by the constraint solver.

There are several parameters and constants in this network.
In the \textproc{ApplyForces} box in the upper-left of figure~\ref{fig:spnet}, $Gravity$ is set to be $-9.8\frac{m}{s^2}$ and ${\Delta}t$ is set to be $\frac{1}{60}$.
The rest density $\rho_0$, shown in the \textproc{ApplyViscosity} box in the lower-right of figure~\ref{fig:spnet} and in the \textproc{SolvePressure} box in figure~\ref{fig:solvepressure}, is set empirically based on the rest density of water.
The fluid parameters $\lambda_\omega$ and $\lambda_v$ are shown in figure~\ref{fig:solvepressure} and the lower-right of figure~\ref{fig:spnet} respectively.
%The fluid parameters $\lambda_c$ and $\lambda_s$ are not shown in the diagram in figure~\ref{fig:net_diagram}, however they are contained in the \textproc{SolveCohesion} and \textproc{SolveSurfaceTension} boxes respectively.

% \begin{figure}
%   \centering
%   \includegraphics[width=13.0cm]{Legend.pdf}
%   \caption{The basic layout for SPNets. TODO}
%   \label{fig:net_diagram}
% \end{figure}

% \begin{figure}
%   \centering
%   \includegraphics[width=10.0cm]{SPNet_diagram.pdf}
%   \caption{The basic layout for SPNets. TODO}
%   \label{fig:net_diagram}
% \end{figure}

% \begin{figure}
%   \centering
%   \includegraphics[width=7.0cm]{ApplyForces_Diagram.pdf}
%   \caption{TODO}
%   \label{fig:net_diagram}
% \end{figure}

% \begin{figure}
%   \centering
%   \includegraphics[width=12.0cm]{ApplyViscosity_Diagram.pdf}
%   \caption{TODO}
%   \label{fig:net_diagram}
% \end{figure}

% \begin{figure}
%   \centering
%   \includegraphics[width=12.0cm]{SolveConstraints_Diagram.pdf}
%   \caption{TODO}
%   \label{fig:net_diagram}
% \end{figure}

% \begin{figure}
%   \centering
%   \includegraphics[width=13.0cm]{SolvePressure_Diagram.pdf}
%   \caption{TODO}
%   \label{fig:net_diagram}
% \end{figure}

% \begin{figure}
%   \centering
%   \includegraphics[width=13.0cm]{SolveObjectCollisions_Diagram.pdf}
%   \caption{TODO}
%   \label{fig:net_diagram}
% \end{figure}

\section{Model Comparison}

We include here a comparison between our model and an established implementation of the same algorithm for verification, which there was not enough room to include in the paper itself.
We compared our model to Nvidia FleX~\cite{macklin2014}, a commercially available physics simulation engine which implements fluid dynamics using Position Based Fluids (PBF).
For this comparison, we set all the model parameters (e.g., the pressure parameter $\lambda_\omega$) to be the same for both FleX and SPNets, we initialize the particle poses and velocities to the same values, and all rigid objects follow the same trajectory.
We compare FleX and SPNets on two scenes.
In the {\it scooping} scene, the liquid rests at the bottom of a large basin as a cup moves in a circle, scooping liquid and then dumping it into the air.
In the {\it ladle}, the liquid rests in a rectangular container as a ladle scoops some from the container and then pours it back into it.
Images from the {\it ladle} scene are shown in figure \ref{fig:ladle}.
We simulate each scene 9 times, once for each combination of values for the cohesion parameter $\lambda_c \in \{0.01, 0.1, 0.2\}$ and viscosity parameter $\lambda_v \in \{0.001, 10, 120\}$ (we fix all other constants).

To compare FleX and SPNets, we compute the intersection over union (IOU) of the two particle sets at each point in time.
To compute the intersection between two particle sets with real valued cartesian coordinates, we relax the ``identical'' constraint to be within a small $\epsilon$, i.e., the intersection is the set of all particles from each set that have at least one neighbor in the opposing set that is within $\epsilon$ units away.
For this comparison we set $\epsilon$ to be 2.5cm.
For the {\it scooping} scene, the IOU was 91.0\%, and for the {\it ladle} scene, the IOU was 97.1\%.
Given this, it is clear that while SPNets is not identical to FleX, it matches closely and produces stable fluid dynamics.

\section{Evaluation Details}

Here we detail how we evaluated our model on the control tasks from section~5.

\subsection{Estimating Fluid Parameters}

To estimate the fluid parameters from data for the results in Section~5.1, we did the following.
We used Nvidia FleX~\cite{macklin2014} to generate ground truth data, and then used backpropagation and gradient descent to iteratively update the parameter values until convergence.
FleX is a commercially available physics simulation engine which implements fluid dynamics using Position Based Fluids (PBF).

Given sequences $\mathcal{P} = \{P_t\}$ and $\mathcal{V} = \{V_t\}$ of particle positions and velocities over time generated by FleX, at each iteration we do the following.
First we randomly sample $B$ particle positions $P$ and velocities $V$ from $\mathcal{P}$,$\mathcal{V}$ to make a training batch $\mathcal{P}_B$, $\mathcal{V}_B$.
Next, SPNet is used to roll out the dynamics $T$ timesteps forward in time to generate $\widetilde{\mathcal{P}}_{B+T}$, $\widetilde{\mathcal{V}}_{B+T}$, the predicted particle positions and velocities after $T$ timesteps.
We then compute the loss $l(\widetilde{\mathcal{P}}_{B+T}, \widetilde{\mathcal{V}}_{B+T}, \mathcal{P}_{B+T}, \mathcal{V}_{B+T})$ between the predicted positions and velocities and the ground truth positions and velocities.
Since our model is differentiable, we can use backpropagation to compute the gradient of the loss with respect to each of the fluid parameters.
We then take a gradient step to update the parameters.
This process is repeated until the parameters converge.

We used the {\it ladle} scene shown in Figure~2a to test our method.
Here, the liquid rests in a rectangular container as a ladle scoops some liquid from the container and then pours it back into the container.
We generated 9 sequences, one for each combination of the cohesion parameter $\lambda_c \in \{0.05, 0.1, 0.15\}$ and viscosity parameter $\lambda_v \in \{30, 60, 90\}$ (we fixed all other fluid parameters).
Each sequence lasted exactly 620 frames.
We set our batch size to 8, $T$ to 2, and use Adam~\cite{kingma2014} with default parameter values and a learning rate of $1\mathrm{e}{-2}$ to update the fluid parameters at each iteration.
We evaluate using 2 different loss functions.
The first is an L1 loss between the predicted and ground truth particle positions and velocities.
This is possible because we know which particle in Flex corresponds to which particle in the SPNet prediction.
In real world settings, however, such a data association is not known, so we evaluate a second loss function that eschews the need for it.
We use the projection loss, which simulates a camera observing the scene and generating binary pixel labels as the observation (similar to the heatmaps generated by our method in chapter~\ref{chapter:perception}).
We compute the projection loss between the predicted and ground truth states by projecting the visible particles onto a virtual camera image, adding a small Gaussian around the projected pixel-positions of each  particle, and then passing the entire image through a sigmoid function to flatten the pixel values.
The loss is then the L1 difference between the projected image of the predicted particles and the ground truth particles.
Projecting the particles as a Gaussian allows us to compute smooth gradients backwards through the projection.
For the {\it ladle} scene, the camera is placed horizontally from the ladle, looking at it from the side.

\subsection{Liquid Control}

\begin{figure}
    \centering
    \includegraphics[width=\linewidth]{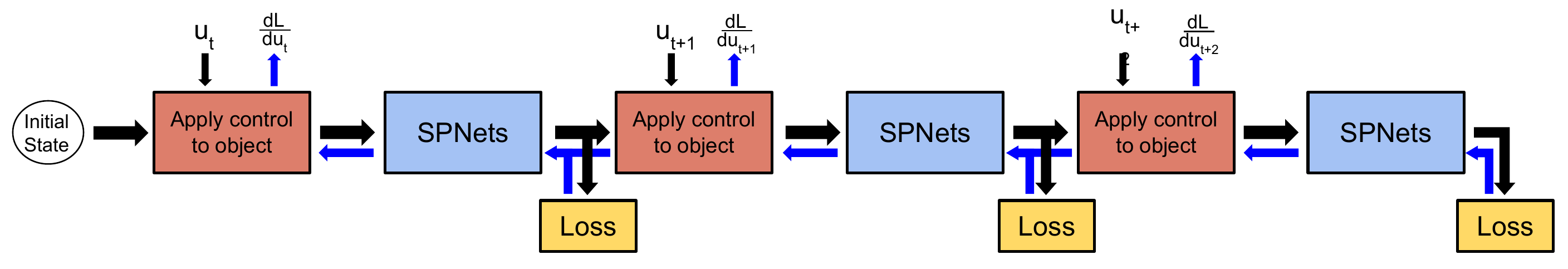}
    \caption{Diagram of the rollout procedure for optimizing the controls $u$. The dynamics are computed forward (black arrows) for a fixed number of timesteps into the future (shown here are 3). Then gradients of the loss are computed with respect to the controls backwards (blue arrows) through the rollout using backpropagation.}
    \label{fig:mpc_rollout}
\end{figure}

For the liquid control results described in Section~5.2, we generated them in the following manner.
The goal in each is to find the set of controls $\mathcal{U} = \{u_t\}$ that minimize the cost
\begin{equation}
L = \displaystyle\sum_t l(P_t, V_t, u_t) \label{eq:mpc_loss}
\end{equation}
where $l$ is the cost function, $P_t$ is the set of particle positions at time $t$, and $V_t$ is the set of particle velocities at time $t$.
$P_t$ and $V_t$ are defined by the dynamics as follows
\begin{equation}
 P_t, V_t = SPN(P_{t-1}, V_{t-1}, OP(u_t)) \label{eq:mpc_dyn}
\end{equation}
where $SPN$ is the fluid dynamics computed by SPNets, and OP transforms the control $u_t$ to the poses of the rigid objects in the scene at time $t$.
The initial positions $P_0$ and velocities $V_0$ of the particles, the loss function $l$, and the control function OP are fixed for each specific control task.

To optimize the controls $\mathcal{U}$, we utilized Model Predictive Control (MPC)~\cite{camacho2013}.
MPC optimizes the controls for a short, finite time horizon, and then re-optimizes at every timestep.
Specifically, given the current particle positions $P_t$ and velocities $V_t$ and the set of controls $\mathcal{U}$ computed at the previous timestep, MPC first computes the future positions $P_{t+1}, ..., P_{t+T}$ and velocities $V_{t+1}, ..., V_{t+T}$ by repeatedly applying the SPNet for some fixed horizon $T$.
Then, MPC sums the loss over this horizon as described in equation~\eqref{eq:mpc_loss} and computes the gradient of the loss $L$ with respect to each control ${\Delta}u_t$ via our differentiable model.
Finally, the updated controls $\mathcal{U}'$ are computed as follows
\[ \mathcal{U}' = \Bigg\{ u_i - s\frac{{\Delta}u_i}{\| {\Delta}u_i \| } \; \Bigg | \; i \in [t, t+T]  \Bigg\} \]
where $s$ is a fixed step size.
The first control $u'_t \in \mathcal{U}'$ is executed, the next particle positions $P_{t+1}$ and velocities $V_{t+1}$ are computed, and this process is repeated to update all controls again.
Note that this process updates not only the current control $u_t$ but also all controls in the horizon, so that by the time control $u_{t+T}$ is executed, it has been updated $T$ times.  
Figure~\ref{fig:mpc_rollout} shows a diagram of this process.
We set $T$ to 10 and use velocity controls on our 3 test scenes. 
The controls $u$ are initialized to 0 and $T$ is set to a fixed horizon for each scene.

We evaluated SPNets on 3 liquid control tasks:
\begin{itemize}

    \item \textbf{The {\it Plate} Scene:} 
Figure~3a shows the {\it plate} scene.
It consists of a plate surrounded by 8 bowls.
A small amount of liquid is placed on the center of the plate, and the plate must be tilted such that the liquid falls into a given target bowl.
The controls for this task are the rotation of the plate about the x (left-right) and z (forward-backward) axes\footnote{In all our scenes, the y axis points up}.
We set the loss function for this scene to be the L2 (i.e., Euclidean) distance between the positions of the particles and a point in the direction of the target bowl.
We ran 8 evaluations on this scene, once with each bowl as the target.

    \item \textbf{The {\it Pouring} Scene:}
We also evaluated our method on the {\it pouring} scene, shown in Figure~3b.
The goal of this task is to pour liquid from the cup into the bowl.
The control is the rotation of the cup about the z (forward-backward) axis, starting from vertical.
Note that there is no limit on the rotation; the cup may rotate freely clockwise or counter-clockwise.
Since the cup needs to perform a two part motion (turning towards the bowl to allow liquid to flow out, then turning back upright to stop the liquid flow), we use a two part piecewise loss function.
For the first part, we set the loss to be the L2 distance between all the liquid particles and a point on the lip of the cup closest to the bowl.
Once a desired amount of liquid has left the cup, we switch to the second part, which is a standard regularization loss, i.e., the loss is the rotation of the cup squared, which encourages it to return upright. We ran 11 evaluations of this scene, varying the desired amount of poured liquid between 75g and 275g.

    \item \textbf{The {\it Catching} Scene:} The final scene we evaluated on was the {\it catching} scene, shown in Figure~3c.
The scene consisted of two cups, a source cup in the air filled with liquid and a target cup on the ground.
The source cup moved arbitrarily while dumping the liquid in a stream.
The goal of this scene is to shift the target cup along the ground to catch the stream of liquid and prevent it from hitting the ground.
The control is the x (left-right) position of the cup.
In order to ensure smooth gradients, we set the loss to be the x distance between each particle and the centerline of the target cup inversely weighted by that particle's distance to the top of the cup.
The source cup always rotated counter-clockwise (CCW), i.e., always poured out its left side.
We ran 8 evaluations of our model, varying the movement of the source cup.
In every case, the source cup would initially move left/right, then after a fixed amount of time, would switch directions.
Half the evaluations started with left movement, the other half right.
We vary the switch time between 3.3s, 4.4s, 5.6s, and 6.7s.

\end{itemize}

\subsection{Learning a Liquid Control Policy via Reinforcement Learning}

\begin{figure}
    \centering
    \includegraphics[width=\linewidth]{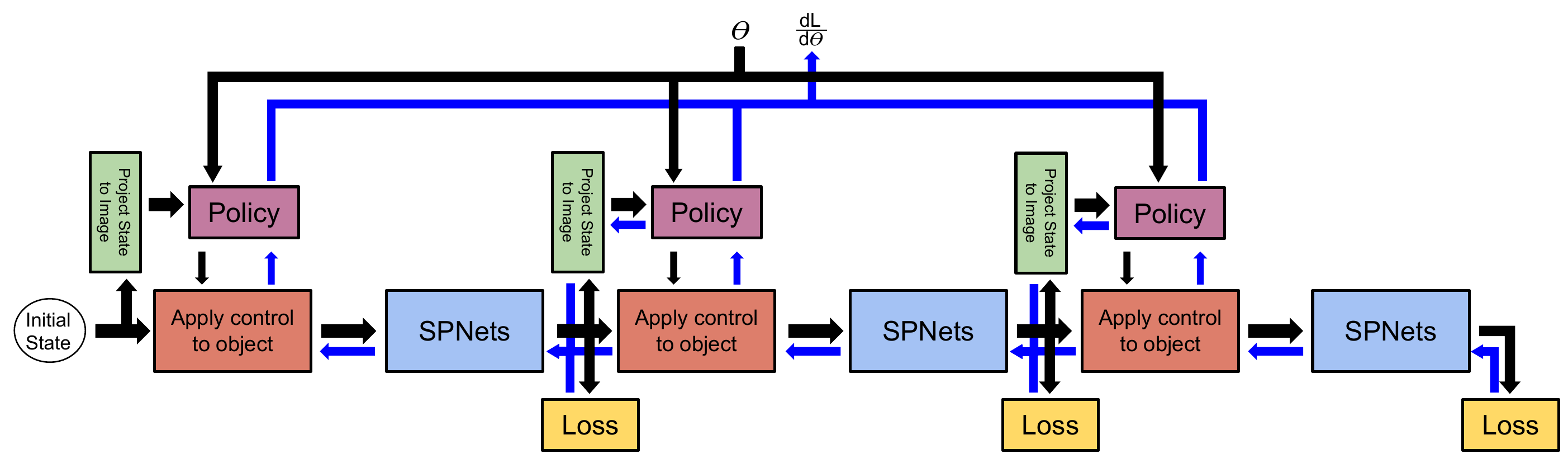}
    \caption{Diagram of the rollout procedure for optimizing the policy parameters $\theta$. This is very similar to the procedure shown in Figure~\ref{fig:mpc_rollout}. The dynamics are computed forward (black arrows) for a fixed number of timesteps into the future (shown here are 3). Then gradients of the loss are computed with respect to $\theta$ backwards (blue arrows) through the rollout using backpropagation.}
    \label{fig:rl_rollout}
\end{figure}

Here we describe the details of how we evaluated our model on the task of learning a policy in a reinforcement learning setting.
Let $u_t$ to be the control at timestep $t$.
It is computed as
\[ u_t = \pi(o_t, \theta) \]
where $o_t$ is the observation at time $t$, $\theta$ are the policy parameters, and $\pi$ is a function mapping the observation (and policy parameters) to controls.
Since we have access to the full state, we compute the observation $o_t$ as a function of the particle positions $P_t$ and velocities $V_t$.
The goal is to learn the parameters $\theta^*$ that best optimize a given loss function $l$.

To do this, we can use a technique very similar to the MPC technique which we described in the previous section.
The main difference is that because the controls $u_t$ are a function of the policy, we optimize instead the policy parameters $\theta$.
We rollout the policy for a fixed number of timesteps, compute the gradient of the policy parameters with respect to the loss, and then update the parameters.
This is possible because our model is fully differentiable, so we can use backpropagation to compute the gradients backwards through the rollout.
Figure~\ref{fig:rl_rollout} shows a diagram of this process.

We test our methodology on the {\it catching} scene.
To train our policy, we use the data generated by the 8 control sequences from the previous section using MPC on the {\it catching} Scene.
At each iteration of training, we randomly sample a different timestep $t$ for each of the 8 sequences, then rollout the policy starting from the particle positions $P_t$ and velocities $V_t$.
We initialize the target cup X position by adding Gaussian noise to the X position of the target cup at time $t$ in the training sequences.
The observation is computed by projecting the particles onto a virtual camera image as described in section~\ref{sec:ex2}.
The camera is positioned so that both cups are in its field of view.
Its X position is set to be the same as the X position of the target cup, that is, the camera moves with the target cup.

Since the observation is effectively binary pixel labels, we use a relatively simple model to learn the policy.
We use a convolutional neural network with 1 convolutional layer (10 $3{\times}3$ kernels with stride of 2) followed by a rectified linear layer, followed by a linear layer with 100 hidden units, followed by another rectified linear layer, and finally a linear layer with 1 hidden unit.
We feed the output through a hyperbolic tangent function to ensure it stays within a fixed range.
We trained the policy for 1200 iterations using the Adam~\cite{kingma2014} optimizer with a learning rate of $2.5\mathrm{e}{-5}$.
The input to the network is a $160{\times}120$ image and the output is the control.

\section{Combining Perception and SPNets}

Here we layout the details of how combined SPNets with perception.
We tasked the robot with tracking the 3D state of liquid over time.
We assume the robot has 3D mesh models of all the rigid objects in the scene and knows their poses at all points in time.
We further assume that the robot knows the initial state of the liquid.
The robot then interacts with the objects and observes the scene with its camera.
The task of the robot is to use its observations from its camera to track the changes in the 3D liquid state over time.
The robot is equipped with an RGB camera for observing the liquid.
We use a thermographic camera aligned to the RGB camera and heated water to acquire ground truth pixel labels.
We refer the reader our prior work \citep{schenckc2018a} for details on the thermographic camera setup.

\subsection{Methodology}

To track the liquid state, the robot takes advantage of its knowledge of fluid dynamics built-in to SPNets.
In an ideal world, knowing the initial state of the liquid and the changes in poses of the rigid objects, it should be possible to simulate the liquid alongside the real liquid, where the simulated liquid would perfectly track the real liquid.
However, no model is perfect and there is inevitably going to be mismatch between the simulation and the real liquid.
Furthermore, due to the temporal nature of this problem, a small error can quickly compound to a large deviation.
The solution in this case is to ``close the loop,'' i.e., use the robot's perception of the real liquid to correct the state of the simulation to prevent errors from compounding and better match the real liquid.

The methodology we adopt here is very similar to that of our prior work~\citep{schenckc2017a} where the robot simulates the liquid forward in time alongside the real liquid while using perception to correct the state at each timestep, however here the robot tracks the liquid state using noisier RGB observations rather than the thermal camera directly.
In this section, we use only pouring interactions, so we simulate each as follows.
Each interaction starts with a known amount of liquid in the source container.
The robot initializes the liquid state by placing a corresponding amount of liquid particles in the 3D mesh of the source container.
Then, at each timestep, the robot updates the poses of the rigid objects and simulates the liquid forward for 1 timestep.
During the simulation step, the robot uses its perception to correct the particle positions (described in the next paragraph).
Each timestep corresponds to $\frac{1}{30}$ of a second in simulation time.
The robot repeats this simulation process until the interaction is over.

The main difference between the methodology here and that in \citep{schenckc2017a} is that instead of assuming the robot has access to an expensive thermographic camera (and is using heated water), we assume the robot has access only to an inexpensive RGB camera.
Thus we must use a different methodology to integrate the perception into the simulation.
In our other prior work~\citep{schenckc2018a} we developed several deep network architectures for producing pixel-wise liquid labels from RGB images.
Here we use the LSTM-FCN with RGB input to convert raw RGB images to pixel-wise liquid labels.
We refer the reader to that paper for more details on that network, which we briefly describe here.
The LSTM-FCN is a fully convolutional neural network that takes in an RGB image and outputs a binary pixel label ({\it liquid} or {\it not-liquid}) for each pixel.
It is recurrent, which means it uses an explicit memory that it passes forward from one timestep to the next (it uses an LSTM layer to enable this recurrence).
It is composed of 6 convolutional layers, each followed by a rectified linear layer.
The first 3 layers are followed by max-pooling layers, the LSTM layer is inserted after the fifth convolutional layer, and the network is terminated with a transposed convolution layer (to upsample the resolution to the original input's size).
We trained the LSTM-FCN using the real robot dataset collected in that paper using the same methodology.
After training the LSTM-FCN, we froze the weights.

We then used the pixel labels output by the LSTM-FCN to correct the state of the simulation.
We treat this perception correction as a constraint, similar to the pressure or cohesion constraints, allowing us to add it to the inner loop of the PBF algorithm (lines 5--9 in Figure~1).
We define the function \textproc{SolvePerception} that takes as input the current set of particle locations $P'$ and the RGB image $R$ and produces ${\Delta}P^R$, the vector to move each particle by to better satisfy the perception constraint.
We insert this function immediately after line~7 and append ${\Delta}P^R$ to the summation on the following line of the algorithm.
This adds the perception correction to be computed alongside the other corrections in the inner-loop of PBF.
Note that because this is added as part of the inner loop in PBF, the velocity is automatically updated based on this correction on line~11.

\begin{figure}
    \centering
    \setlength{\fboxsep}{0pt}
    \setlength{\fboxrule}{1pt}
    \setlength{\unitlength}{1.0cm}
    \includegraphics[width=\linewidth]{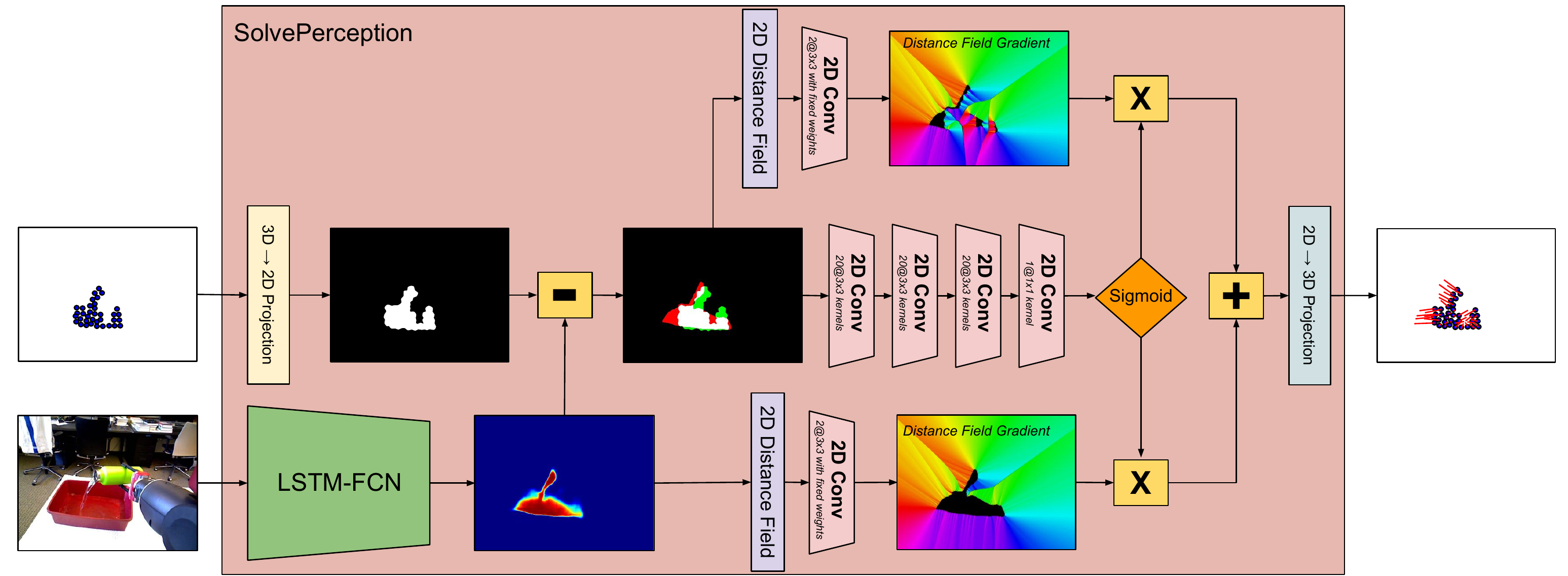}
    \caption{A diagram of the \textproc{SolvePerception} method. It takes as input the current particle state (center-left) and the current RGB image (lower-left). The RGB image is passed through the LSTM-FCN to generate pixel labels, the particle state is projected onto 2D. From this, 2 gradient fields are computed, where the gradient points in the direction of the closest {\it liquid} pixel. These fields are then blended and projected onto the particles in 3D.}
    \label{fig:spnets_perception_diagram}
\end{figure}

Figure~\ref{fig:spnets_perception_diagram} shows a diagram of how we implement the \textproc{SolvePerception} function.
We first apply the LSTM-FCN to the RGB image to generate pixel-wise liquid labels, which we refer to as {\it observed} labels.
Next we compute the {\it observed} distance field over the {\it observed} labels, i.e., the distance from each pixel to the closest positive liquid pixel.
We use this and a 2D convolution with fixed parameters to compute the {\it observed} distance field gradient, i.e., for every pixel the direction to the closest positive liquid pixel.
In parallel, we project the state of the particles onto a 2D image using the camera's intrinsic and extrinsic parameters.
This also generates a pixel-wise label image, which we refer to as the {\it model} labels.
We then subtract the {\it model} labels from the {\it observed} labels to generate the {\it disparity} image, i.e., the pixels for which the {\it model} and {\it observation} disagree.
We then generate the {\it disparity} distance field gradient in the same manner as for the {\it observed} distance field gradient.
Furthermore, we feed the {\it disparity} image through 4 2D convolutions (the first three of which are followed by a ReLU) and a sigmoid function.
The output of this is a blending value for each pixel.
These blending values determine how to combine the two distance field gradients.
We multiply the {\it observed} distance field gradient by the blending values, and the {\it disparity} distance field gradient by 1 minus the blending values and add them together.
This results in a blended gradient field. 
We project this back onto the particles in 3D, adjusting the gradient by the camera parameters.
The result is the set ${\Delta}P^R$, the distance to move each particle to better match the perception.
Note that for both projections (projecting 3D particles onto the 2D image plane and projecting the 2D gradient field onto the 3D particles), we ignore particles that are blocked from view of the camera by an object (e.g., particles in a container).

\begin{figure}
    \centering
    \includegraphics[width=\linewidth]{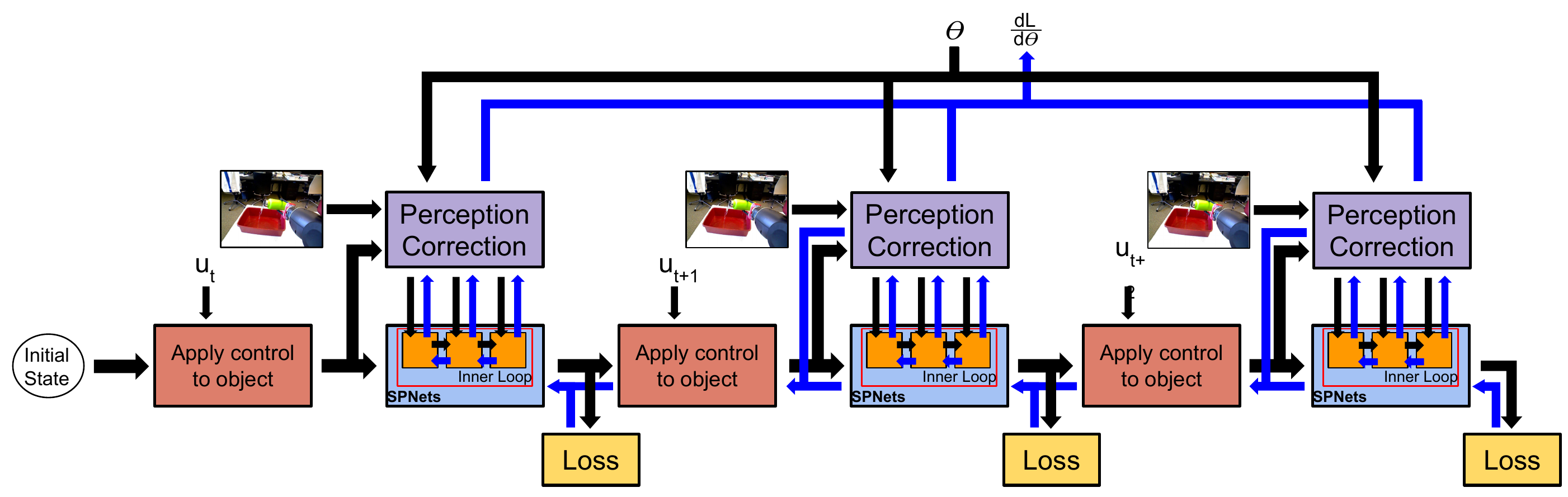}
    \caption{Diagram of the rollout procedure for optimizing the parameters of the perception network $\theta$. This is very similar to the procedures shown in Figures \ref{fig:mpc_rollout} and \ref{fig:rl_rollout}. The dynamics are computed forward (black arrows) for a fixed number of timesteps into the future (shown here are 3). In this case the controls $u$ are fixed. The gradients of the loss are computed with respect to $\theta$ backwards (blue arrows) through the rollout using backpropagation. The box ``Perception Correction'' corresponds to the network shown in Figure~\ref{fig:spnets_perception_diagram} and $\theta$ are its parameters. Note that here the inner loop of the PBF algorithm in SPNets is shown (with 3 iterations) since this is how the perception correction is intergrated into the dynamics.}
    \label{fig:perception_rollout}
\end{figure}

To train the parameters of the network, we do the following.
We train the parameters of the LSTM-FCN part of the network using the same training data and methodology as in our prior work~\citep{schenckc2018a} and we refer the reader there for details.
We then fix those parameters for the remainder of the training.
We pre-train the entire network in Figure~\ref{fig:spnets_perception_diagram} by randomly selecting frames from our dataset and randomly placing particles in the scene.
To compute the loss, we use the ground truth pixel labels collected from the thermal camera as described in our prior work.
The loss is computed as
\[ L(\mathcal{L}, P) = \left( \displaystyle\sum_{p \in P} \displaystyle\min_{l \in \mathcal{L}} || PROJ(p) - l ||^2 \right) + \left( \displaystyle\sum_{l \in \mathcal{L}} \displaystyle\min_{p \in P} || l - PROJ(p) ||^2 \right) \]
where $\mathcal{L}$ is the set of positive liquid pixels in the ground truth image, $P$ is the set of particle positions, and $PROJ$ projects a particle location from 3D onto the 2D image plane.
Intuitively, this loss computes 2 terms: the {\it accuracy}, i.e., how far each particle is from a liquid pixel, and the {\it coverage}, i.e., how far each liquid pixel is from a particle or how well the particles cover the liquid pixels.
We pre-train this network for 48,000 iterations using ADAM \cite{kingma2014} with a learning rate of 0.0001, default momentum values, and a batch size of 4.
Finally, we train the network from end-to-end by adding \textproc{SolvePerception} into SPNets, unrolling it over time, and computing the same loss.
Again, this is possible because SPNets can propagate the gradient backwards in time from one timestep to the previous, allowing us to use those gradients to update the learned weights.
We trained the network this way for 3,500 iterations also using ADAM with the same learning rate and momentum values, a batch size of 1, and unrolling for 30 timesteps\footnote{Unrolling the network for training here is the same unrolling technique we used to train the LSTM-FCN in \citep{schenckc2018a}.}.
A diagram showing this training process is shown in Figure~\ref{fig:perception_rollout}.

\subsection{Evaluation}

We evaluated SPNets combined with perception on the 12 {\it pouring} sequences we collected on the real robot.
That is, the robot executed 12 pouring sequences following a fixed trajectory with 2 different source containers (a cup and a bottle).
One third of the sequences the container started 30\% full, one third 60\% full, and one third 90\% full.
We tracked the liquid state using both SPNets with perception and SPNets alone for comparison.
For each sequence, the known amount of liquid was placed in the source container at the start, and as the object poses were updated at each timestep, the liquid was also updated via SPNets.
For SPNets with perception, \textproc{SolvePerception} was added to the PBF algorithm as described in the previous section and the RGB image from the robot's camera was used at each timestep as input to that function.
For SPNets alone, \textproc{SolvePerception} was not added and the liquid state was tracked open-loop instead.

We evaluate the intersection-over-union (IOU) across all frames of the 12 pouring sequences.
To compute the IOU, we compare the {\it true} pixel labels from the ground truth (gathered using the thermal camera) with the {\it model} pixel labels.
To get the {\it model} pixel labels, we project each particle in the simulation onto the 2D image plane.
However, since there are far fewer particles than pixels, we draw a circle of radius 5 around each particle's projected location.
The result is a set of pixel labels corresponding to the state of the model.
To compute the IOU, we divide the number of pixels where the {\it model} and {\it true} labels agree by the number of pixels that are positive in either set of labels.